%% file: Deepfake.tex
\documentclass[sigconf]{acmart}
\usepackage{multirow}
\usepackage{float}
\usepackage{enumitem}
\usepackage{amsmath}
\usepackage{algorithmic}

\usepackage[ruled, linesnumbered]{algorithm2e}
\usepackage{xr}
\usepackage[table]{xcolor}
\usepackage{colortbl}
\usepackage{subcaption}
\usepackage{pifont}
\usepackage{makecell}
\theoremstyle{remark}

\setcopyright{none}

\begin{document}

% \title{Null‑space Regularization and Intervention:
 % From Training to Inference for Robust Deepfake Detection}

\title{Suppressing Forgery-Specific Shortcuts for Generalizable Deepfake Detection}

% % \renewcommand{\thefootnote}{\fnsymbol{footnote}}
% \footnotetext[1]{These authors contributed equally to this work.}
% \footnotetext[2]{Corresponding author.}

% Yihui Wang, Yonghui Yang, Jilong Liu, Fengbin Zhu, Le Wu, Tat-Seng Chua

\author{Yihui Wang}
\affiliation{
% \department{Key Laboratory of Knowledge Engineering with Big Data,}
\institution{Hefei University of Technology}
% \city{Hefei}
\country{China}
}
\email{MistryNihilityn@gmail.com}

\author{Yonghui Yang}
\authornotemark[1]
\affiliation{
% \department{Key Laboratory of Knowledge Engineering with Big Data,}
\institution{National University of Singapore}
% \city{Hefei}
\country{Singapore}
}
\email{yyh.hfut@gmail.com}
\thanks{Yonghui Yang is the Corresponding author}

\author{Jilong Liu}
\affiliation{
% \department{Key Laboratory of Knowledge Engineering with Big Data,}
\institution{Hefei University of Technology}
% \city{Hefei}
\country{China}
}
\email{liujilong0116@gmail.com}

\author{Fengbin Zhu}
\affiliation{
% \department{Key Laboratory of Knowledge Engineering with Big Data,}
\institution{National University of Singapore}
% \city{Hefei}
\country{Singapore}
}
\email{fengbin@nus.edu.sg}

\author{Le Wu}
\affiliation{
% \department{Key Laboratory of Knowledge Engineering with Big Data,}
\institution{Hefei University of Technology}
% \city{Hefei}
\country{China}
}
\email{lewu.ustc@gmail.com}

\author{Tat-Seng Chua}
\affiliation{
% \department{Key Laboratory of Knowledge Engineering with Big Data,}
\institution{National University of Singapore}
% \city{Hefei}
\country{Singapore}
}
\email{dcscts@nus.edu.sg}

\renewcommand{\shortauthors}{Yihui Wang et al.}

\newcommand{\fullname}{\textit{\textbf{S}hortcut-\textbf{S}ubspace \textbf{S}uppression~(\textbf{S$^3$})}}
\newcommand{\shortname}{S$^3$}

\input{Deepfake_Sections/0-abs}

% \begin{CCSXML}
% <ccs2012>
%    <concept>
%        <concept_id>10002951.10003317.10003347.10003350</concept_id>
%        <concept_desc>Information systems~Recommender systems</concept_desc>
%        <concept_significance>500</concept_significance>
%        </concept>
%  </ccs2012>
% \end{CCSXML}

% \ccsdesc[500]{Information systems~Recommender systems}


\maketitle

\input{Deepfake_Sections/1.1-intro}

\input{Deepfake_Sections/3-method}

\input{Deepfake_Sections/4-experiments}

\input{Deepfake_Sections/2-related-works}

\input{Deepfake_Sections/5-Conclusion}

% \clearpage
\bibliographystyle{ACM-Reference-Format}
\balance
\bibliography{Deepfake}

\clearpage
\appendix

\input{Deepfake_Sections/6-Appendix}

\end{document}

%% file: Deepfake_Sections/0-abs.tex
\begin{abstract}

Deepfake detection suffers from poor generalization across forgery methods, as existing models tend to rely on spurious method-specific shortcuts that fail to transfer to unseen manipulations. While recent approaches attempt to improve generalization, they lack an explicit mechanism to identify and suppress such shortcuts in learned representations. In this work, we propose \fullname~framework that explicitly characterizes and suppresses method-specific shortcuts via subspace modeling. Our key insight is that variations distinguishing different forgery methods capture method-specific artifacts and thus serve as an effective proxy for method-specific shortcuts. To this end, we train a lightweight linear probe for forgery method classification and perform Singular Value Decomposition (SVD) to extract the dominant shortcut subspace. Building on this formulation, we develop two complementary strategies to reduce shortcut reliance. During training, we softly suppress the shortcut subspace in feature representations, encouraging the model to rely on more generalizable cues for real/fake discrimination. At inference time, we introduce a training-free counterpart that attenuates neurons aligned with the identified shortcut directions, enabling plug-and-play generalization enhancement with improved interpretability. Extensive experiments on multiple benchmarks demonstrate that our method significantly improves cross-method generalization while maintaining strong in-domain performance. The code will be released upon acceptance of the submission.
\end{abstract}

%% file: Deepfake_Sections/1.1-intro.tex
\section{Introduction}
Recent advances in deep generative models, particularly GANs~\cite{goodfellow2014generative} and diffusion models~\cite{ho2020denoising,rombach2022high}, have significantly improved the realism and diversity of synthetic visual content, attracting widespread attention in both academia and industry. While these techniques enable powerful applications in content creation, they also raise serious concerns, including misinformation, identity impersonation, and potential misuse in political contexts~\cite{tolosana2020deepfakes, mirsky2021creation}. As a result, deepfake detection has become an increasingly important task in real-world applications. Consequently, a wide range of detection methods have been proposed~\cite{roessler2019faceforensicspp, afchar2018mesonet, masi2020two, qian2020thinking, wang2023dire, yermakov2025unlocking} to distinguish manipulated content from authentic images. Despite their promising performance, these methods often struggle to maintain reliability when encountering unseen forgery methods~\cite{ojha2023towards, yan2023ucf}, limiting their applicability in practical scenarios.

Such generalization failure is fundamentally rooted in the open-world nature of deepfake detection, where the forgery distribution is non-stationary and cannot be exhaustively covered during training~\cite{yao2023towards}. As a result, although existing detectors achieve strong performance on known manipulation types, they often fail to generalize to unseen forgeries. Researchers have explored several directions to improve detection generalization, mainly following the principles: (i) leveraging more transferable forgery cues~\cite{durall2020watch,liu2021spatial,  tan2024frequency}, (ii) expanding the diversity of forgery patterns through data augmentation or synthesis~\cite{shiohara2022detecting, yan2024transcending}, and (iii) learning invariant representations to reduce reliance on method-specific biases~\cite{yin2024improving, dong2023implicit, yan2024orthogonal}. While these approaches have improved robustness to some extent, the underlying mechanisms that govern generalization behavior remain not fully understood. In particular, existing methods often lack an explicit understanding of which components of the learned representations are responsible for generalization across forgery types. Consequently, models may inadvertently rely on spurious cues that do not transfer well to unseen forgeries. This suggests the need for a more principled perspective on representation-level generalization in deepfake detection.

\begin{figure}[t]
\vspace{-0.8em}
    \centering 
  \subfloat[Xception]{
      \includegraphics[width=0.225\textwidth]{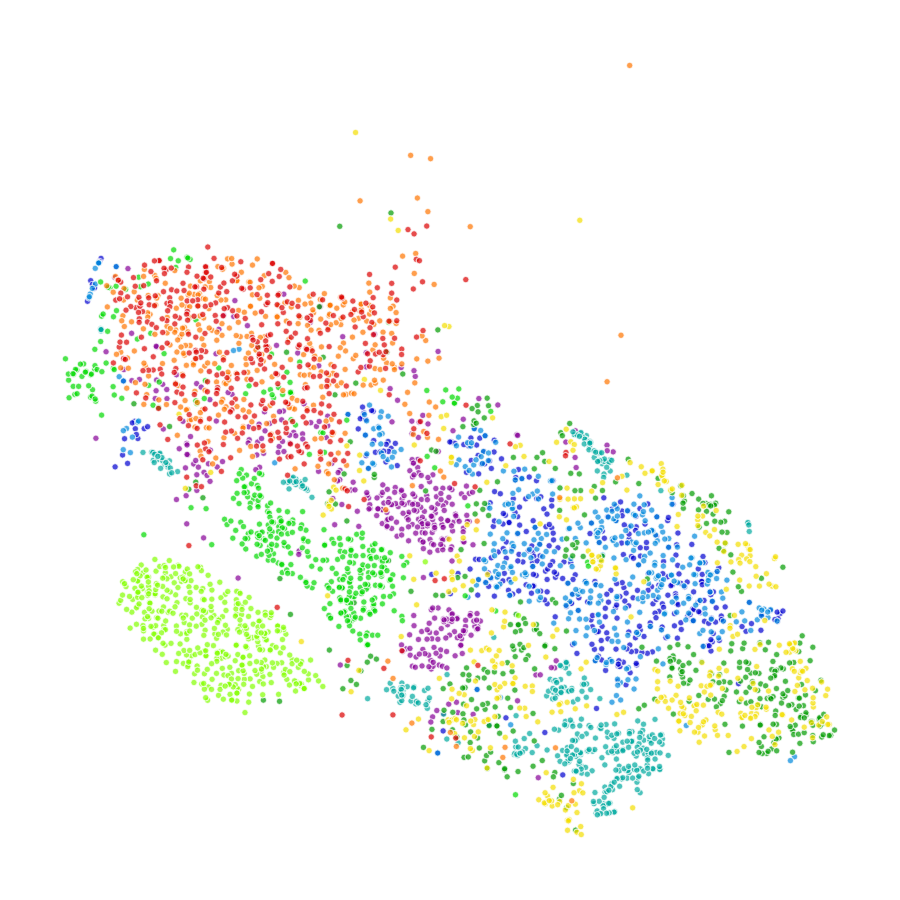}
      \label{fig:baseline-cluster}
  }
  \subfloat[Projected]{
      \includegraphics[width=0.225\textwidth]{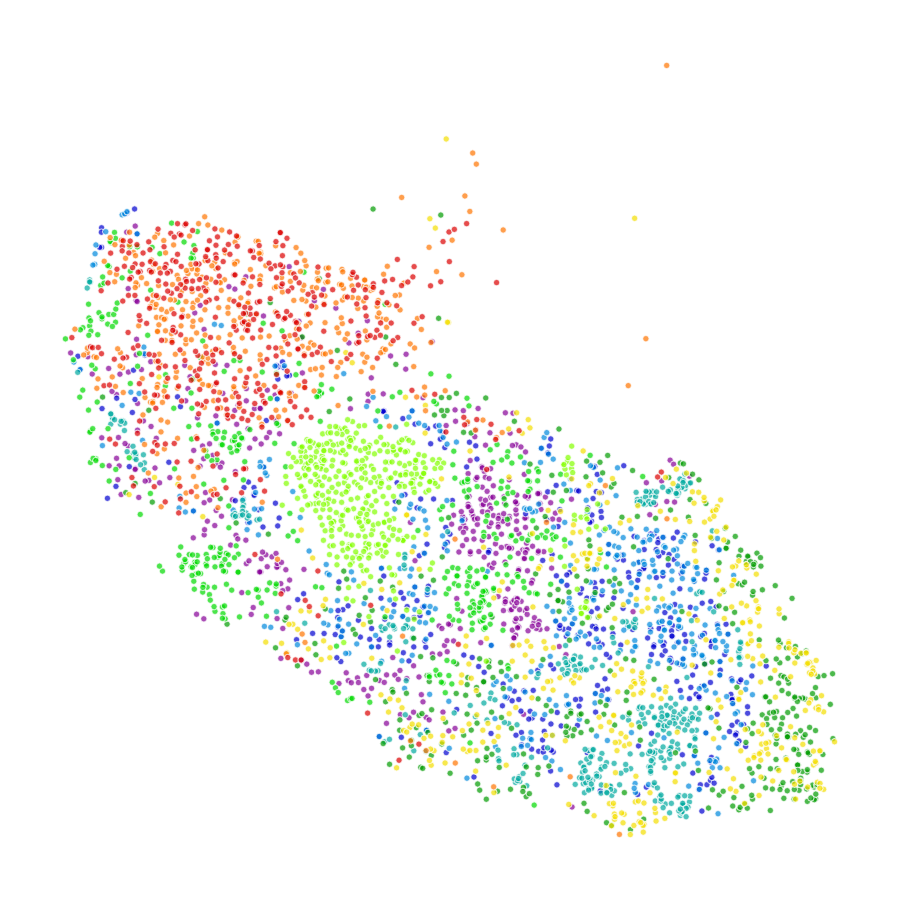}
      \label{fig:nullspace-cluster}
  }

    \subfloat{
      \includegraphics[width=0.45\textwidth]{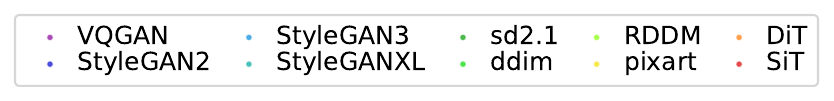}
      \label{fig:color-legend}
  }
  \setlength{\belowcaptionskip}{-0.5cm}
  \caption{t-SNE visualization on DF40 (EFS). (a) Xception Baseline: real (red) and fake (other colors) are clearly separable; fake samples colored by forgery method form distinct clusters, revealing method-specific representation patterns. (b) After projecting features onto the nullspace of the method-sensitive subspace, method-specific clusters dissolve while real/fake separation persists.}
  
  \label{fig:tnse-EFS-baseline-nullspace} 
\end{figure}

In this work, we revisit deepfake detection from a representation-level perspective, aiming to better understand how different components of learned features contribute to generalization across forgery methods. We first conduct empirical analyses and visualize the feature distributions across different forgery methods. As shown in Figure ~\ref{fig:tnse-EFS-baseline-nullspace} (a), fake samples form distinct, well-separated clusters according to their forgery technique. This indicates that the model's representations are dominated by surface textures or patterns unique to each forgery method, rather than by unified discriminative cues. Although learned representations exhibit clear method-dependent clustering patterns, the entanglement with real/fake discrimination signals makes it challenging to isolate the method-specific components.

\begin{figure}[t]
  \centering 
  \subfloat[Energy distribution]{
      \includegraphics[width=0.225\textwidth]{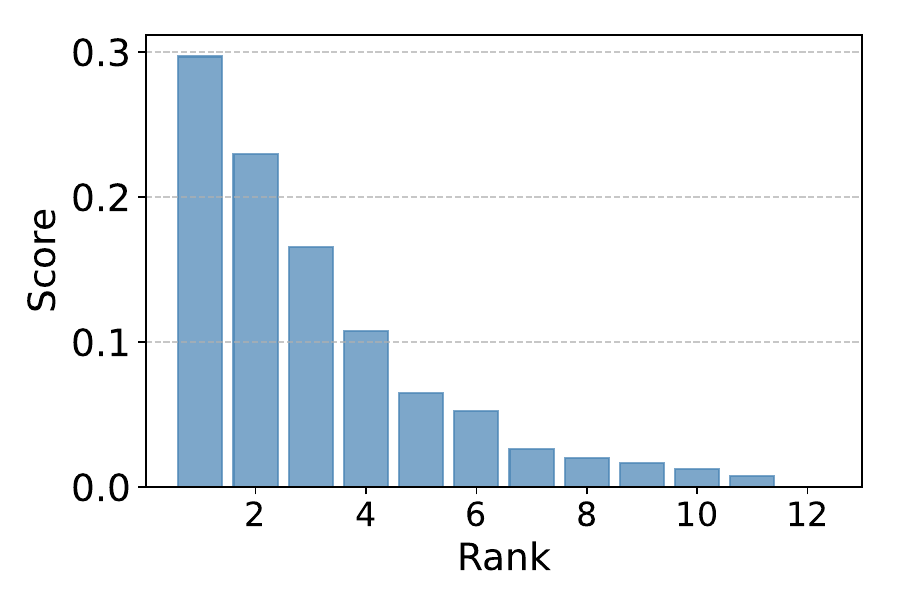}
      \label{fig:singular-energy}
  }
  \subfloat[Detection performance]{
      \includegraphics[width=0.225\textwidth]{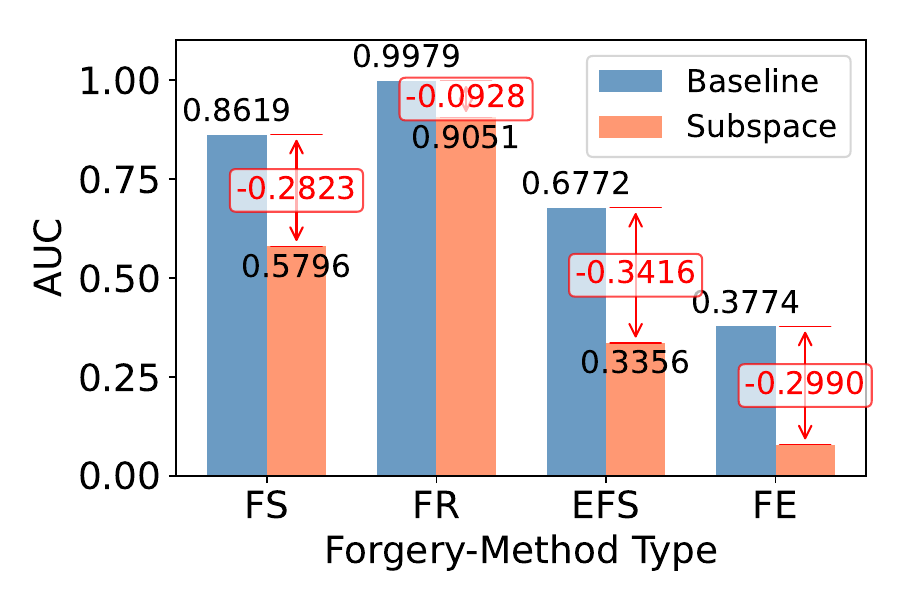}
      \label{fig:subspace-results}
  }
  \setlength{\belowcaptionskip}{-0.5cm}
  \caption{(a) Singular value energy distribution: the first few components concentrate most energy, indicating a low-rank method-sensitive subspace. (b) AUC comparison between the original and subspace-projected features input. On the training domain (FR), the projected features retains most performance; on unseen domains (FS, EFS, FE), its AUC collapses, confirming that the subspace encodes the shortcut directions in the feature space.}
  \label{fig:energy-subspace} 
\end{figure}

To better understand and explicitly characterize the method-specific components, we learn the method-sensitive subspace in the feature space. Specifically, we train a lightweight probe to predict forgery methods from learned representations, thereby extracting method-discriminative signals encoded in the features. We then perform SVD on the probe to identify the principal directions that are most sensitive to forgery-method variations, which span the method-sensitive subspace. This subspace captures the dominant directions for distinguishing different forgery methods and typically exhibits a low-rank structure, as shown in Figure~\ref{fig:singular-energy}. Through this formulation, method-specific features become explicitly analyzable, enabling us to directly study their role in model behavior. Specifically, we project the learned representations onto the method-sensitive subspace and then use the projected features for real/fake classification. As illustrated in Figure~\ref{fig:subspace-results}, we compare the detection performances between the original and projected features. We find that the projected features retains most performance on the training domain, but exhibits significant collapses on unseen domains. This indicates that the method-sensitive subspace is the shortcut directions in the feature space, which directly cause the generalization failure to unseen forgery methods.

Building on the above formulation, we propose a unified \fullname~framework that mitigates method-specific shortcuts through two complementary strategies. During training, we introduce a \textbf{nullspace-based projection mechanism} that explicitly suppresses the gradient components aligned with the method-sensitive subspace, thereby preventing the model from updating along shortcut directions. This encourages the detector to focus on more transferable cues for real/fake discrimination, rather than overfitting to method-specific patterns~(as shown in Figure~\ref{fig:nullspace-cluster}). At inference time, we further analyze neuron activation patterns with respect to both forgery-method discrimination and real/fake prediction. Based on this analysis, we identify \textbf{shortcut neurons} that are highly responsive to forgery-method variations while contributing greatly to the real/fake decision. We then suppress their activations without modifying model parameters, resulting in a backbone-training-free, plug-and-play intervention that directly reduces shortcut influence at test time. These two strategies provide a unified framework that mitigates method-specific shortcuts from both the optimization and inference stages, leading to improved cross-method generalization with enhanced interpretability. Our contributions are summarized as follows:
\begin{itemize}[leftmargin=2.0em]
    \item We provide a systematic empirical analysis revealing that generalization failure in deepfake detection stems from method-specific shortcuts, i.e., non-transferable patterns learned during training.
    \item We explicitly characterize these shortcuts as a method-sensitive subspace in the feature space, offering a principled formulation for understanding shortcut behavior.
    \item We propose two complementary suppression strategies, including a training-stage nullspace projection and an inference-stage shortcut neuron activation editing, which together improve generalization with limited computational overhead.
    \item Extensive experiments demonstrate that our method consistently outperforms state-of-the-art approaches across multiple benchmarks, achieving superior cross-method generalization while maintaining strong in-domain performance. 
\end{itemize}

%% file: Deepfake_Sections/3-method.tex
\section{Preliminary}
In this section, we briefly formalize the deepfake detection problem and the objective of cross-method generalization.

\paragraph{Problem Setup.}
\label{sec:problem-setup}
Let $x \in \mathcal{X}$ denote an input image and $y \in \{0,1\}$ its label, where $y=0$ indicates a real image and $y=1$ a fake one. Each fake sample is further associated with a forgery method label $m \in \mathcal{M}$, where $\mathcal{M}$ denotes the set of forgery methods observed during training. A common deepfake detector consists of a feature extractor $\phi_\theta: \mathcal{X} \rightarrow \mathbb{R}^d$(i.e., ResNet~\cite{He_2016_CVPR}, Xception~\cite{Chollet_2017_CVPR}, CLIP~\cite{pmlr-v139-radford21a}) and a classifier $\psi: \mathbb{R}^d \rightarrow [0,1]$, yielding:
\begin{equation}
f(x) = \psi(\phi_\theta(x)).
\end{equation}
During training, supervision is available for both $y$ and $m$, while at test time, the model is evaluated on samples whose forgery methods may lie outside $\mathcal{M}$. 
% This unseen forgery led to the generalization failure of current detectors

\paragraph{Generalization Objective.}
Deepfake detection is inherently an open-world problem, where the distribution of forgery methods is non-stationary. Let $\mathcal{D}_{\text{train}}$ and $\mathcal{D}_{\text{test}}$ denote the training and testing distributions, respectively, with potentially disjoint method sets. The generalization objective is to learn the optimal representation $z = \phi_\theta(x)$ on $\mathcal{D}_{\text{train}}$, which can achieve the minimal test risk:
\begin{equation}
\mathbb{E}_{(x,y)\sim \mathcal{D}_{\text{test}}}
\big[ \ell(\psi(z), y) \big].
\end{equation}
In practice, the learned representation often captures signals that are highly predictive on $\mathcal{D}_{\text{train}}$ but hard to transfer across different forgery methods~\cite{yan2023ucf, ojha2023towards, yao2023towards}. This drives us to explore more generalizable deepfake detection.

\section{Methodology}

\begin{figure*}[t]
    \centering
    \includegraphics[width=0.95\textwidth]{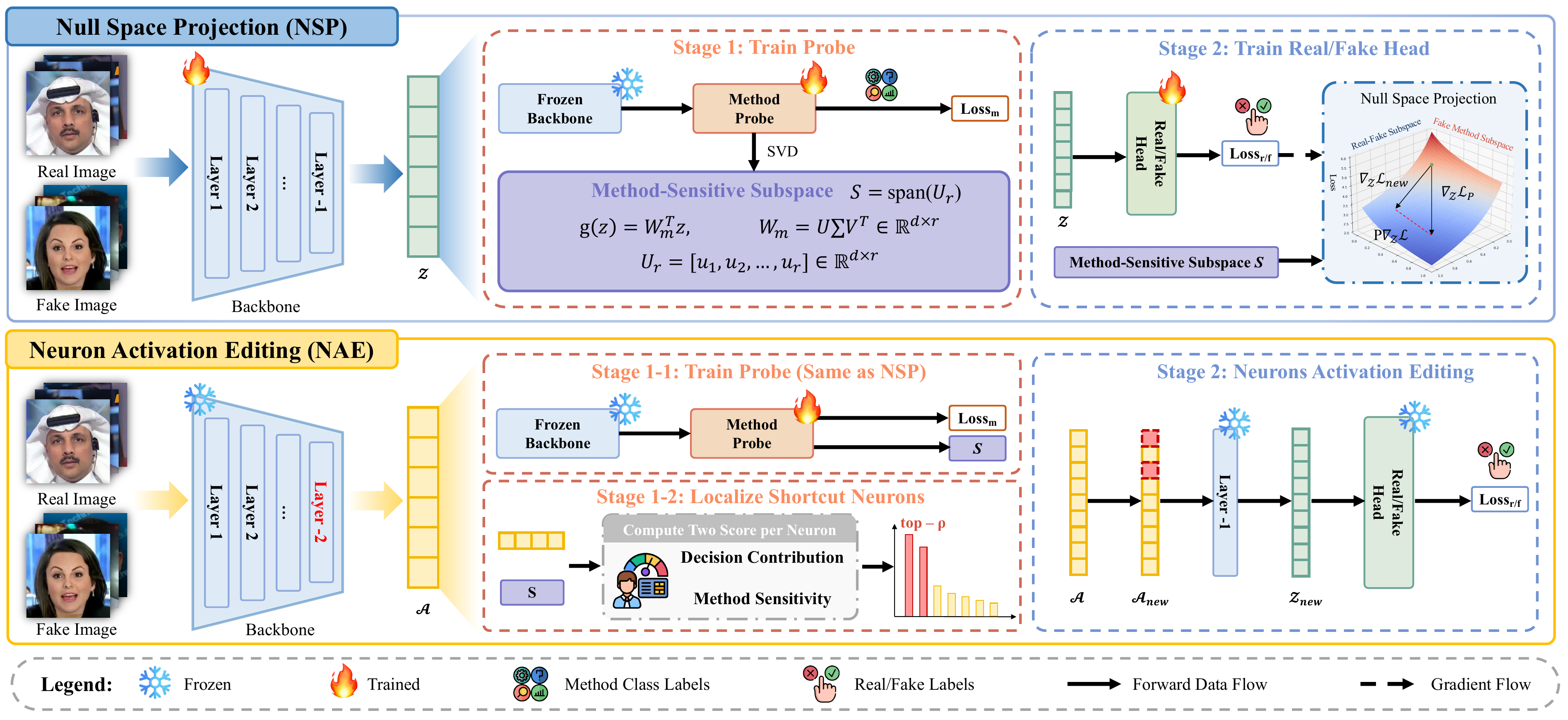}
    \caption{Overview of the proposed \shortname~framework. The upper part illustrates training-time subspace suppression (NSP) with alternating probe training and gradient projection; the lower part shows inference-time neuron activation editing (NAE) with probe training and activation editing rules.}
    \label{fig:s3_overview}
\end{figure*}

In this work, we attribute the generalization failure to the method-specific shortcut. To tackle the shortcut issue, we first design an explicit shortcut formulation mechanism by method-sensitive subspace extraction. Next, we introduce our proposed \fullname~framework illustrated in Figure~\ref{fig:s3_overview} for generalizable deepfake detection, which involves two complementary shortcut suppression strategies during training and inference stages, respectively. Finally, we give a comprehensive discussion about our proposed \shortname~framework.

\subsection{Method-Sensitive Subspace}
To explicitly characterize method-specific shortcuts in the feature space, we begin by analyzing how forgery method information is encoded in learned representations. Given a representation $\mathbf{z} = \phi_\theta(x)$, we consider predicting the forgery method $m$ from $\mathbf{z}$. If certain directions in the feature space are highly predictive of $m$, they capture variations that distinguish different method pipelines. Since such variations are inherently tied to specific manipulation methods, they are unlikely to generalize across unseen methods and thus constitute shortcut signals. To identify these directions, we introduce a linear probe:
\begin{equation}
g(\mathbf{z}) = W_m^\top \mathbf{z},
\end{equation}
where $W_m \in \mathbb{R}^{d \times K}$ and $K = |\mathcal{M}|$. We adopt a linear probe for two reasons. 
First, linear models expose directly which directions in the representation space are predictive of the target, providing interpretability~\cite{alain2016understanding, bau2017network, kim2018interpretability}. 
Second, empirical evidence suggests that high-level semantic attributes are often linearly separable in deep representations, making linear probes a standard tool for analyzing learned features across both vision and language models~\cite{harkonen2020ganspace, belinkov2022probing}. 

The weight matrix $W_m$ encodes how each feature dimension contributes to distinguishing different forgery methods. To extract the dominant method-predictive directions, we perform Singular Value Decomposition (SVD):
\begin{equation}
W_m = U \Sigma V^\top.
\label{eq:svd}
\end{equation}
The left singular vectors $U$ define orthogonal directions in the feature space, ordered by their importance in explaining method-discriminative variations. Empirically, we observe that a small number of singular directions explain most of the method-discriminative power, indicating that shortcut signals reside in a low-dimensional subspace (as shown in Figure 2(a)). We then select the top-$r$ singular vectors to form an orthonormal basis $U_r \in \mathbb{R}^{d \times r}$, which spans a low-dimensional subspace:
\begin{equation}
\mathcal{S} = \mathrm{span}(U_r).
\end{equation}
This subspace captures the principal directions along which different forgery methods are distinguished, and therefore serves as a proxy for method-specific shortcuts in the representation space. Based on this formulation, we propose \fullname~ framework from two complementary perspectives. During training, we constrain gradient updates to avoid this subspace; at inference, we attenuate neurons aligned with these directions. We next introduce these two strategies in detail.

\subsection{Training-time Subspace Suppression}
\label{sec:nsp}

Based on the identified method-sensitive subspace $\mathcal{S}$, we first introduce a training-time strategy to suppress shortcut reliance by directly constraining the optimization dynamics. During training, model parameters are updated along the gradient direction of the loss. If the gradient contains a significant component aligned with the method-sensitive subspace $\mathcal{S}$, the model will increasingly rely on shortcut features, reinforcing method-specific patterns that do not generalize. To mitigate this effect, we propose \textbf{Null Space Projection (NSP)}, which explicitly attenuates gradient components lying in $\mathcal{S}$. Instead of allowing unrestricted updates in the full feature space, NSP constrains the optimization trajectory to avoid shortcut directions, encouraging the model to learn more transferable representations.

Consider the feature vectors $\mathbf{Z}$ extracted from the image backbone and let $\nabla_{\mathbf{Z}} \mathcal{L}$ be the gradient of the loss with respect to $\mathbf{Z}$. Given the orthonormal basis $U_r$ of the method-sensitive subspace, we construct the projection matrix:
\begin{equation}
\mathbf{P} = U_r U_r^\top,
\end{equation}
which projects any vector onto $\mathcal{S}$, and its complement $\mathbf{P}^\perp = \mathbf{I} - \mathbf{P}$, which projects onto the null space of $\mathcal{S}$. We can decompose the gradient into two components:
\begin{equation}
\nabla_{\mathcal{Z}} \mathcal{L} = \mathbf{P} \nabla_{\mathcal{Z}} \mathcal{L} + \mathbf{P}^\perp \nabla_{\mathcal{Z}} \mathcal{L}
\end{equation}
corresponding to the shortcut-aligned and shortcut-orthogonal directions, respectively. Thus, NSP suppresses the shortcut-aligned component by applying:
\begin{equation}
\nabla_{\mathcal{Z}} \mathcal{L}_{\text{new}} 
= \nabla_{\mathcal{Z}} \mathcal{L} - \alpha \mathbf{P} \nabla_{\mathcal{Z}} \mathcal{L}
= \mathbf{P}^\perp \nabla_{\mathcal{Z}} \mathcal{L} + (1 - \alpha)\mathbf{P} \nabla_{\mathcal{Z}} \mathcal{L},
\label{eq:nsp_proj}
\end{equation}
where $\alpha \in [0,1]$ controls the suppression strength. When $\alpha = 1$, the gradient component within $\mathcal{S}$ is completely removed, and the update lies entirely in the null space of the shortcut subspace. For $0 < \alpha < 1$, NSP performs soft suppression, retaining partial information while discouraging over-reliance on shortcut directions. 

By restricting gradient updates along method-sensitive directions, NSP prevents the model from further amplifying shortcut features during training. As a result, the learned representation is biased toward components that are less dependent on specific forgery methods and thus more likely to generalize across unseen forgery methods.

\begin{figure}[t]
  \centering 
  \subfloat[] {
      \includegraphics[width=0.225\textwidth]{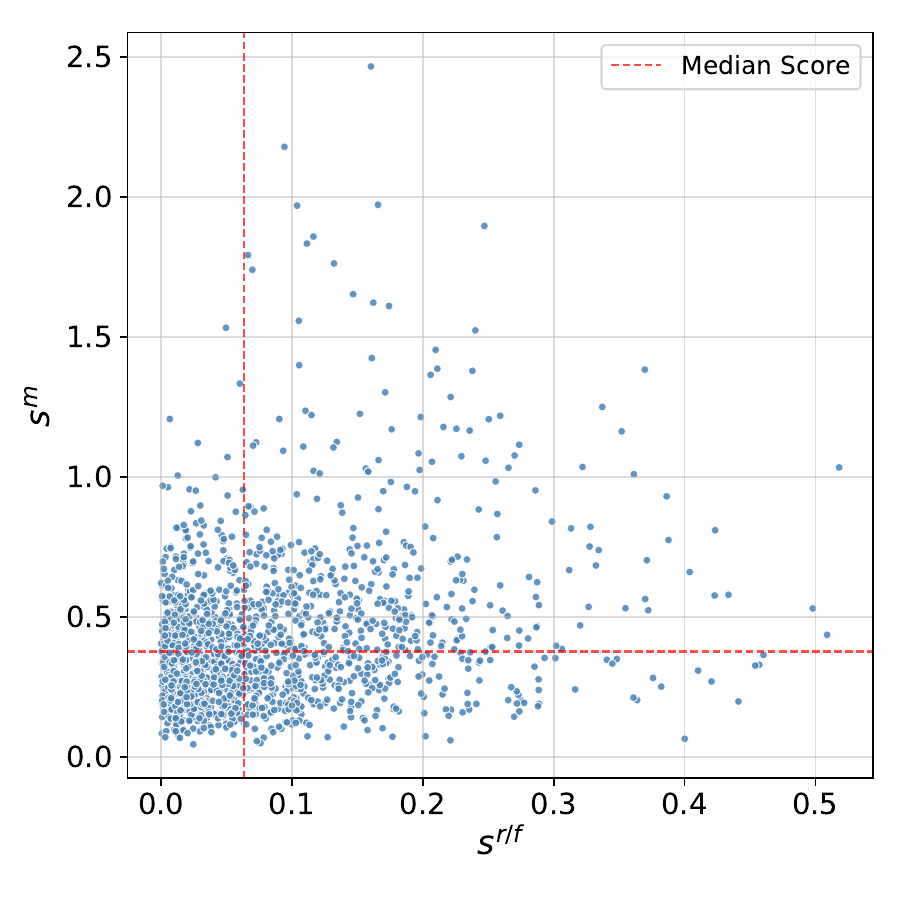}
      \label{fig:correlation-scatter}
  }
  \subfloat[]{
      \includegraphics[width=0.225\textwidth]{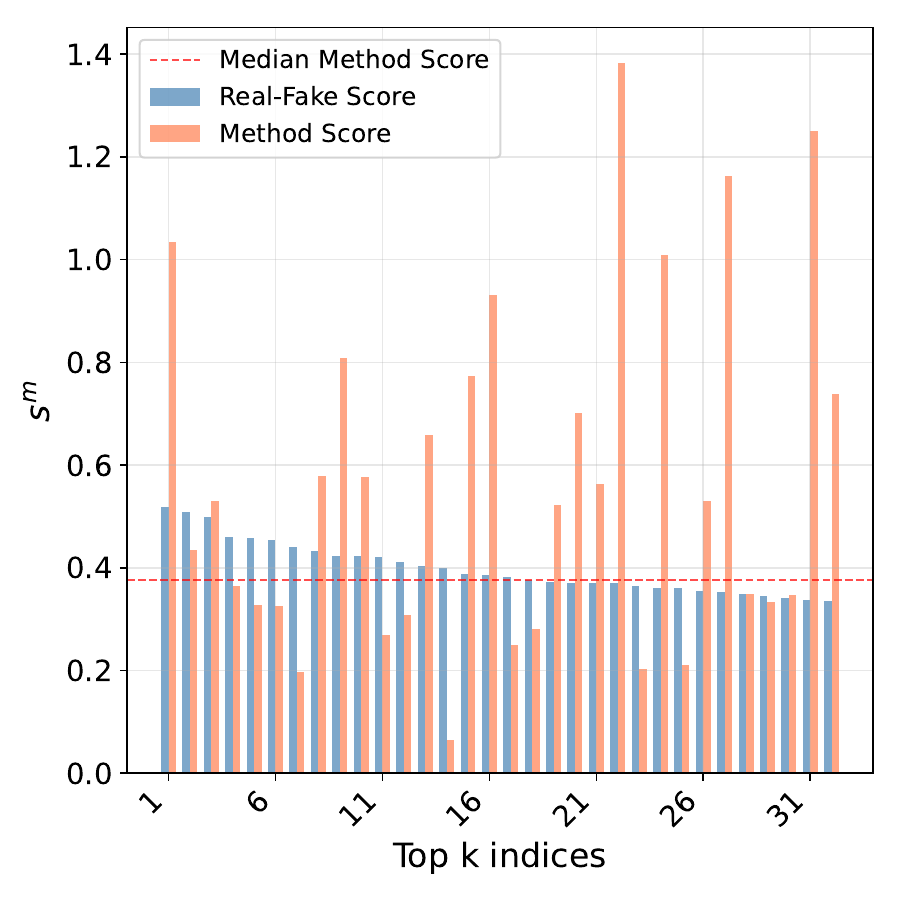}
      \label{fig:top-k-comparison}
  }
  \caption{(a) Neuron Distribution: Decision contribution $s_i^{\text{r/f}}$ vs. method sensitivity $s_i^{\text{m}}$. (b) Top-$k$ decision-critical neurons: many also exhibit high method sensitivity.}
  \label{fig:combine-scores} 
\end{figure}

\subsection{Inference-time Neuron Suppression}
\label{sec:nae}

While NSP suppresses shortcut learning during training, we further propose an inference-time strategy to mitigate shortcut reliance in already trained models. 
Unlike NSP, this approach is training-free and does not require modifying model parameters, making it a practical plug-and-play solution.
Given the method-sensitive subspace $\mathcal{S}$ identified in Sec.~3.1, we analyze how shortcut signals manifest at the neuron level. 
Since neuron activations form the intermediate representation driving the final prediction, neurons whose activations are strongly aligned with $\mathcal{S}$ capture method-specific variations. 
Meanwhile, neurons contribute differently to the real/fake decision. 
This leads to a unified characterization: neurons that are both \emph{decision-critical} and \emph{method-sensitive} are likely to encode shortcut signals. 
Such a formulation provides an interpretable view of the model, allowing us to explicitly localize shortcut-related neurons and enable targeted intervention.

% \paragraph{Quantifying Neuron Contributions.}
To quantify neuron contributions, we adopt a local linear approximation. For a properly chosen \emph{localization layer} (See Appendix~\ref{app:arch_details}), the model's decision logit can be approximated as:
\begin{equation}
\mathbf{n}_{\text{r/f}}^\top \mathbf{z} = \mathbf{c}^\top \mathbf{a} + \text{const},
\end{equation}
where $\mathbf{n}_{\text{r/f}}$ is the normal vector of the decision boundary, $\mathbf{a}$ denotes neuron activations and $\mathbf{c}$ measures their contribution to the final prediction. 
This formulation implies that scaling neuron activations directly modulates the decision outcome, providing a principled basis for intervention. We characterize each neuron using two complementary scores. For a neuron with weight vector $\mathbf{w}_i$, its \emph{decision contribution} is defined as:
\begin{equation}
s_i^{\text{r/f}} = \frac{|\mathbf{w}_i \cdot \mathbf{n}_{\text{r/f}}|}{\|\mathbf{w}_i\| \|\mathbf{n}_{\text{r/f}}\|},
\label{eq:srf}
\end{equation}
which measures alignment with the decision boundary. Its \emph{method sensitivity} is defined as:
\begin{equation}
s_i^{\text{m}} = \sum_{j=1}^{r} \left| (\mathbf{U}_r^{\mathsf{T}} \mathbf{w}_i)_j \right| \cdot \sigma_j,
\label{eq:sm}
\end{equation}
which measures alignment with the method-sensitive subspace $\mathcal{S}$. Together, these scores provide an interpretable decomposition of neuron roles.

% \paragraph{Shortcut Neuron Localization.}
We visualize neurons in a 2D plane defined by $(s_i^{\text{r/f}}, s_i^{\text{m}})$, revealing a clear structure (Fig.~\ref{fig:combine-scores}). 
Notably, neurons that are both highly decision-critical and strongly method-sensitive form a distinct group, which we identify as \emph{shortcut neurons}. 
These neurons dominate the prediction while encoding method-specific signals, making them the primary source of generalization failure.
% \paragraph{Neuron Activation Editing (NAE).}

Based on this analysis, we propose \textbf{Neuron Activation Editing (NAE)}, an inference-time intervention that suppresses shortcut neurons without modifying model parameters.
We first select neurons that are both:
(i) above-median in decision contribution, and  
(ii) among the top-$\rho$ fraction in method sensitivity.
Let $\mathcal{I}$ denote the selected neuron set. 
During inference, we attenuate their activations:
\begin{equation}
\mathcal{A}_{\text{new}}[:, i] = \mathcal{A}[:, i] \cdot (1 - \alpha), \quad \forall i \in \mathcal{I},
\label{eq:nae}
\end{equation}
where $\alpha \in [0,1]$ controls the suppression strength. NAE offers two key advantages. First, it provides an interpretable mechanism for identifying and mitigating shortcut reliance at the neuron level. 
Second, it is entirely training-free and does not require modifying backbone parameters, making it directly applicable to existing models. The full operation process of the proposed \fullname~framework is illustrated in Algorithm~\ref{alg:s3} (Appendix).

\subsection{Discussion of \shortname~Framework}

% \paragraph{Interpretable Mechanism.}
\textbf{Interpretable Mechanism.} 
Our key insight is that method-specific shortcuts occupy a low-dimensional subspace that is complementary to transferable forensic cues. By explicitly identifying and suppressing this subspace, our framework unifies training-time and inference-time interventions under a common perspective. This subspace-based formulation not only improves cross-method generalization, but also provides an interpretable mechanism for understanding and mitigating shortcut reliance. 

% \paragraph{Complexity and Generality.}
\textbf{Simplicity and Generality.}
NSP introduces only a lightweight projection during optimization, while NAE is entirely training-free and can be directly applied to pre-trained models without modifying backbone parameters. Both operations incur negligible overhead and scale independently of model size, making the framework efficient in real-world deployment. Furthermore, our proposed framework operates directly on feature representations and requires minimal assumptions about model architecture, making it compatible with various detectors.

%% file: Deepfake_Sections/4-experiments.tex
\section{Experiments}
We evaluate NSP and NAE from three complementary perspectives. First, we assess the generalization of the cross-method by training on a single forgery type and testing in four distinct categories. Second, we examine the plug‑and‑play nature of our methods by integrating them into four backbone architectures. Third, we analyze computational efficiency to demonstrate practical deployability. To understand \textit{why} our methods work, we further conduct feature‑space visualizations and neuron‑level sensitivity analysis.

\subsection{Experimental Setup}

\subsubsection{Datasets}
We perform experiments on the DF40 dataset~\cite{NEURIPS2024_34239f60}, a large‑scale benchmark comprising 40 deepfake manipulation techniques across four categories: \textbf{Face Swapping (FS)}, \textbf{Face Reenactment (FR)}, \textbf{Complete Face Synthesis (EFS)} and \textbf{Face Editing (FE)}. Each category comprises multiple specific manipulation techniques. See Table~\ref{tab:fake_methods} for the full list. Following the standard protocol, we use the FF domain, where authentic data originates from FaceForensics++~\cite{roessler2019faceforensicspp} with the official split of 720 videos for training and 140 for validation/testing.

To evaluate generalization under distribution shift, we construct three training sets, each containing a single category of forgery methods from the FF domain. For testing, we evaluate all models on the FF domain across all four forgery categories, covering 32 distinct manipulation methods. This setup allows us to measure performance when training and testing forgery types are (i) identical (in‑domain), (ii) distinct (cross‑domain).

\subsubsection{Baselines}
We compare against twelve representative models, spanning both fundamental feature extractors and specialized forensic detectors. This includes four widely adopted backbones—Xception~\cite{Chollet_2017_CVPR}, ResNet50~\cite{He_2016_CVPR}, EfficientNet‑B4\cite{pmlr-v97-tan19a}, and CLIP (ViT‑B/16)~\cite{pmlr-v139-radford21a}—as well as eight state‑of‑the‑art deepfake detectors: F$^3$Net~\cite{qian2020thinking}, SPSL~\cite{liu2021spatial}, SRM~\cite{Lee_2019_ICCV}, RECCE~\cite{Cao_2022_CVPR}, UCF~\cite{yan2023ucf}, FreqNet~\cite{tan2024frequency}, Deepspace~\cite{Roy2025DeepSpaceNT}, and SpecXNet~\cite{SpecXNet}. This diverse selection provides a thorough benchmark for assessing both the effectiveness and generalization capability of our approach.

\subsubsection{Evaluation Metric}
Following prior work, we adopt the Area Under the Curve (AUC) as the primary evaluation metric.

\subsubsection{Implementation Details}
Our implementation builds on DeepfakeBench~\cite{yan2023deepfakebench}, with adaptations to accommodate our proposed modules. All experiments run on a single NVIDIA RTX 4090 GPU (24 GB memory) using PyTorch. Unless constrained by memory or convergence, we use a unified configuration: batch size 128, backbone initialized with pre‑trained weights, and the Adam optimizer with a warmup schedule followed by cosine annealing for learning rate decay.

\subsection{Main Results}
\begin{table*}[ht]
    \centering
    \caption{Performance comparison of different methods under various training sets}
    \tabcolsep=0.1cm
    \label{tab:training_sets}
    \resizebox{0.95\textwidth}{!}{ 
    \begin{tabular}{r|ccccc|ccccc|ccccc}
    \toprule[1.2pt]
    \textbf{Training Set} & \multicolumn{5}{c}{\textbf{FS(FF)}} & \multicolumn{5}{|c}{\textbf{FR(FF)}} & \multicolumn{5}{|c}{\textbf{EFS(FF)}} \\
    \midrule
    \midrule
    \textbf{Methods} & \textbf{FS} & \textbf{FR} & \textbf{EFS} & \textbf{FE} &\textbf{Avg.} & \textbf{FS} & \textbf{FR} & \textbf{EFS} & \textbf{FE} & \textbf{Avg.} & \textbf{FS} & \textbf{FR} & \textbf{EFS} & \textbf{FE} & \textbf{Avg.}  \\
    \midrule
    Xception (ICCV2019)&  0.9832 & 0.8708 & 0.8615 & \underline{0.9801} & 0.9239 & 0.8619 & \textbf{0.9979} & 0.6772 & 0.3774 & 0.7286 & 0.7697 & 0.8221 & \underline{0.9988} & 0.9912 & 0.8955 \\
    F$^3$Net (ECCV2020)& 0.9487 & 0.7482 & 0.7648 & 0.8508 & 0.8281 & 0.7267 & 0.9686 & 0.6422 & 0.2935 & 0.6577 & 0.6651 & 0.7251 & 0.9742 & 0.9763 & 0.8352 \\
    SPSL (CVPR2021)& 0.9772 & 0.8545 & 0.7834 & 0.9666 & 0.8954 & 0.8527 & 0.9957 & 0.6656 & 0.7354 & 0.8123 & 0.7761 & \underline{0.8615} & 0.9965 & 0.9896 & 0.9059 \\
    SRM (CVPR2021)& 0.9713 & 0.8497 & 0.8117 & 0.9586 & 0.8978 & 0.8489 & 0.9972 & 0.6932 & 0.6953 & 0.8087 & 0.7906 & \textbf{0.8878} & 0.9976 & 0.9923 & \textbf{0.9171} \\
    RECCE (CVPR2022)& 0.8637 & 0.7250 & 0.6894 & 0.8943 & 0.7931 & 0.8087 & 0.9837 & 0.6893 & 0.5174 & 0.7498 & 0.6987 & 0.7068 & 0.9450 & 0.9498 & 0.8251 \\
    UCF (ICCV2023)& 0.9766 & 0.8413 & 0.8632 & 0.9558 & 0.9092 & 0.8512 & 0.9961 & 0.7028 & 0.4003 & 0.7376 & 0.6896 & 0.7279 & 0.9969 & 0.9947 & 0.8523 \\
    FreqNet (AAAI2024)& 0.8458 & \textbf{0.9389} & 0.6306 & 0.8990 & 0.8286 & 0.8520 & 0.9904 & 0.7161 & \textbf{0.9217} & \underline{0.8701} & 0.5940 & 0.6845 & 0.8955 & 0.8659 & 0.7600 \\
    Deepspace (VISIGRAPP2025)& 0.9839 & 0.8911 & 0.8487 & 0.9587 & 0.9206 & \underline{0.8698} & \underline{0.9977} & 0.7060 & 0.5144 & 0.7720 & 0.7686 & 0.8120 & \textbf{0.9989} & 0.9961 & 0.8939 \\
    SpecXNet (ACMMM2025)& 0.8956 & 0.6762 & 0.6303 & 0.8317 & 0.7584 & 0.6897 & 0.9597 & 0.5200 & 0.2893 & 0.6147 & 0.6397 & 0.6152 & 0.9731 & 0.9779 & 0.8015 \\
    \midrule
    \shortname-NSP (Ours)& \textbf{0.9858} & \underline{0.9067} & \textbf{0.8841} & 0.9686 & \textbf{0.9363} & 0.8692 & 0.9974 & \underline{0.7609} & 0.6738 & 0.8253 & \textbf{0.8021} & 0.8352 & 0.9985 & \textbf{0.9988} & \underline{0.9087} \\
    \shortname-NAE (Ours)& 0.9840 & 0.8711 & 0.8611 & \textbf{0.9817} & 0.9245 & \textbf{0.8761} & 0.9973 & \textbf{0.8366} & \underline{0.9066} & \textbf{0.9042} & 0.7668 & 0.8293 & 0.9985 & 0.9912 & 0.8965 \\
    \shortname-NSP+NAE (Ours)& \underline{0.9856} & 0.9060 & \underline{0.8837} & 0.9679 & \underline{0.9358} & 0.8464 & 0.9916 & 0.7403 & 0.7636 & 0.8355 & \underline{0.7967} & 0.8351 & 0.9981 & \underline{0.9983} & 0.9071 \\
    \bottomrule[1.2pt]
    \end{tabular}}
\end{table*}
\begin{table*}[ht]
\centering
\caption{Generalization analysis of different models under various training and testing forgery types.}
\label{tab:generalization}
\renewcommand{\arraystretch}{0.5}
\resizebox{0.95\textwidth}{!}{%
\begin{tabular}{l l cccccccccccc}
\toprule
\multirow{2}{*}{\textbf{Backbone}} & \multirow{2}{*}{\textbf{Variant}} & \multicolumn{4}{c}{FS train} & \multicolumn{4}{c}{FR train} & \multicolumn{4}{c}{EFS train} \\
\cmidrule(lr){3-6} \cmidrule(lr){7-10} \cmidrule(lr){11-14}
& & FS test & FR test & EFS test & FE test & FS test & FR test & EFS test & FE test & FS test & FR test & EFS test & FE test \\
\midrule
\multirow{4}{*}{Xception} 
    & baseline & 0.9832 & 0.8708 & 0.8615 & 0.9801 & 0.8619 & \textbf{0.9979} & 0.6772 & 0.3774 & 0.7697 & 0.8221 & \textbf{0.9988} & 0.9912 \\
    & +NSP     & \textbf{0.9858} & \textbf{0.9067} & \textbf{0.8841} & 0.9686 & 0.8692 & 0.9974 & 0.7609 & 0.6738 & \textbf{0.8021} & \textbf{0.8352} & 0.9985 & \textbf{0.9988} \\
    & +NAE     & 0.9840 & 0.8711 & 0.8611 & \textbf{0.9817} & \textbf{0.8761} & 0.9973 & \textbf{0.8366} & \textbf{0.9066} & 0.7668 & 0.8293 & 0.9985 & 0.9912 \\
    & +NSP+NAE & 0.9856 & 0.9060 & 0.8837 & 0.9679 & 0.8464 & 0.9916 & 0.7403 & 0.7636 & 0.7967 & 0.8351 & 0.9981 & 0.9983 \\
\midrule

\multirow{4}{*}{ResNet} 
    & baseline & 0.9892 & 0.9093 & 0.8592 & 0.9617 & 0.8601 & 0.9985 & 0.6767 & 0.4025 & 0.7601 & 0.8391 & 0.9992 & \textbf{0.9983} \\
    & +NSP     & 0.9890 & 0.9089 & 0.8626 & 0.9606 & 0.8840 & 0.9989 & 0.7604 & 0.6047 & \textbf{0.7853} & 0.8392 & \textbf{0.9996} & 0.9959 \\
    & +NAE     & \textbf{0.9894} & 0.9103 & 0.8610 & \textbf{0.9621} & 0.8582 & 0.9985 & 0.7001 & 0.4061 & 0.7605 & \textbf{0.8415} & 0.9992 & 0.9982 \\
    & +NSP+NAE & 0.9888 & \textbf{0.9162} & \textbf{0.8695} & 0.9533 & \textbf{0.8850} & \textbf{0.9990} & \textbf{0.7784} & \textbf{0.6395} & 0.7609 & 0.8209 & \textbf{0.9996} & 0.9953 \\
\midrule
\multirow{4}{*}{EfficientNet}
    & baseline & \textbf{0.9867} & 0.9136 & 0.8657 & 0.9803 & 0.8901 & 0.9997 & 0.5992 & 0.7184 & 0.7982 & 0.8634 & 0.9998 & 0.9980 \\
    & +NSP     & 0.9840 & \textbf{0.9264} & \textbf{0.8748} & \textbf{0.9942} & 0.9012 & \textbf{0.9998} & 0.6288 & 0.7455 & 0.8249 & 0.8918 & \textbf{0.9999} & \textbf{0.9998} \\
    & +NAE     & 0.9866 & 0.9108 & 0.8632 & 0.9793 & 0.8890 & 0.9996 & 0.6273 & 0.7171 & 0.8006 & 0.8658 & 0.9998 & 0.9977 \\
    & +NSP+NAE & 0.9838 & 0.9214 & 0.8730 & 0.9940 & \textbf{0.9015} & \textbf{0.9998} & \textbf{0.6330} & \textbf{0.7532} & \textbf{0.8338} & \textbf{0.8955} & \textbf{0.9999} & 0.9996 \\
\midrule
\multirow{4}{*}{CLIP}
    & baseline & 0.9804 & 0.8398 & 0.7921 & 0.8806 & 0.8742 & \textbf{0.9990} & 0.7145 & 0.6707 & 0.7469 & 0.8682 & \textbf{0.9980} & \textbf{0.9997} \\
    & +NSP     & 0.9836 & 0.8723 & 0.8095 & 0.9049 & 0.8683 & 0.9988 & \textbf{0.7244} & 0.6994 & 0.7526 & 0.8590 & \textbf{0.9980} & 0.9991 \\
    & +NAE     & 0.9806 & 0.8584 & 0.8000 & 0.8967 & \textbf{0.8765} & \textbf{0.9990} & 0.7165 & 0.6866 & \textbf{0.7542} & \textbf{0.8747} & \textbf{0.9980} & \textbf{0.9997} \\
    & +NSP+NAE & \textbf{0.9839} & \textbf{0.8798} & \textbf{0.8201} & \textbf{0.9237} & 0.8677 & 0.9988 & \textbf{0.7244} & \textbf{0.7030} & 0.7522 & 0.8596 & \textbf{0.9980} & 0.9991 \\
\bottomrule
\end{tabular}}
\end{table*}

\subsubsection{Quantitative Comparison}

When training and test forgery types diverge, baseline detectors often collapse—but our methods consistently recover performance. Table~\ref{tab:training_sets} compares NSP and NAE against twelve baselines under three single‑forgery training regimes (FS, FR, EFS). The key observation: cross‑domain performance gains are largest precisely where baselines struggle most. For instance, training on FS and testing on FR—a distribution shift that drops Xception from 0.9832 (in‑domain FS) to 0.8708—NSP raises this to 0.9067, a 4.12\% relative improvement. Similarly, NAE proves especially effective when baselines catastrophically fail: trained on FR and tested on FE, Xception achieves only 0.3774; NAE boosts this to 0.9066, a 140\% gain. These results directly validate our core hypothesis: suppressing method‑specific shortcuts—whether during training or at inference—forces the model to rely on transferable forensic cues.

\subsubsection{Generalization across Backbones}
\label{sec:generalizable_exp}

The effectiveness of NSP and NAE extends beyond Xception to diverse architectures. Table~\ref{tab:generalization} shows results across ResNet50, EfficientNet‑B4, and CLIP (ViT‑B/16), with three consistent patterns.

\textbf{Architecture‑agnostic gains.} Both modules improve cross‑domain performance across all backbones, with the most dramatic gains under severe distribution shifts. For ResNet50 trained on FR and tested on EFS, NSP improves AUC from 0.6767 to 0.7604 (+12.4\%); while NAE achieves a modest gain to 0.7001 (+3.5\%). These improvements are not Xception specific---they reflect a general mechanism for decoupling method‑specific shortcuts from decision‑critical features.

\textbf{Complementary roles.} 
NSP acts as a proactive regularizer, consistently improving cross-domain performance across most settings. NAE, by contrast, is a post-hoc remedy which typically yields smaller gains than NSP. When combined, NSP+NAE does not always outperform either alone, as shown in Tables~\ref{tab:training_sets} and \ref{tab:generalization}. For instance, on FE the AUC drops from 0.9066 with NAE alone to 0.7636 with the combination when trained on FR, suggesting the two mechanisms suppress overlapping shortcut directions and cause over‑attenuation. Under EFS training, however, they complement each other, raising the FE AUC from 0.9912 with NAE alone to 0.9983 with the combination. This domain‑dependent interaction shows that training‑time and inference‑time suppression are not always additive; their compatibility hinges on how strongly the model relies on method‑specific shortcuts.

\textbf{Modest in‑domain trade‑off.} When training and test forgery types match, baseline AUCs already exceed 0.98. Adding NSP or NAE yields minor fluctuations ($\leq$ 0.3\% decrease). This slight in‑domain cost is expected: suppressing method‑specific features intentionally sacrifices capacity to exploit patterns that are effective only within the training distribution. The resulting cross‑domain gains—often exceeding 10 $\times$ the in‑domain loss—demonstrate a favorable trade‑off for real‑world deployment where forgery methods are rarely known in advance.

\subsubsection{Efficiency Analysis}
\label{sec:efficiency}
Our methods add minimal computational overhead. NSP introduces 30 seconds per epoch (42\% increase) with no extra GPU memory. NAE requires a one‑time probe training of 16 seconds (1.74 GB). At inference, both add <0.12 ms per image and 18.5K parameters (0.09\% of the backbone). In contrast, competing detectors like SRM and UCF exceed 24 GB memory under the same batch size, and SpecXNet is 9× slower. Table~\ref{tab:efficiency} summarizes the key comparisons; the full results and architecture‑specific details are deferred to Appendix~\ref{app:efficiency}.

\begin{table}[ht]
\centering
\caption{Efficiency comparison (abridged). Full results in Appendix~\ref{app:efficiency}.}
\label{tab:efficiency}
\resizebox{0.45\textwidth}{!}{\begin{tabular}{lcccc}
\toprule
\textbf{Method} & \makecell{Train time \\ (s/epoch)} & \makecell{GPU mem \\ (GB)} & \makecell{Infer time \\ (ms/img)} & \makecell{Params} \\
\midrule
Xception (baseline) & 72 & 12.27 & 5.57 & 20.8M \\
+ NSP (ours) & 73+30& 12.27 & 5.64 & +18.5K \\
+ NAE (ours) & 0+16 & 1.74 & 5.69 & +18.5K \\
\midrule
SRM & — & >24 & 10.22 & 53.2M \\
SpecXNet & 330 & 16.57 & 51.70 & 53.8M \\
\bottomrule
\end{tabular}}
\end{table}

\subsection{Ablation Study}
\label{sec:ablation}

\begin{figure*}[!t]
  \centering 
  \includegraphics[width=0.92\textwidth]{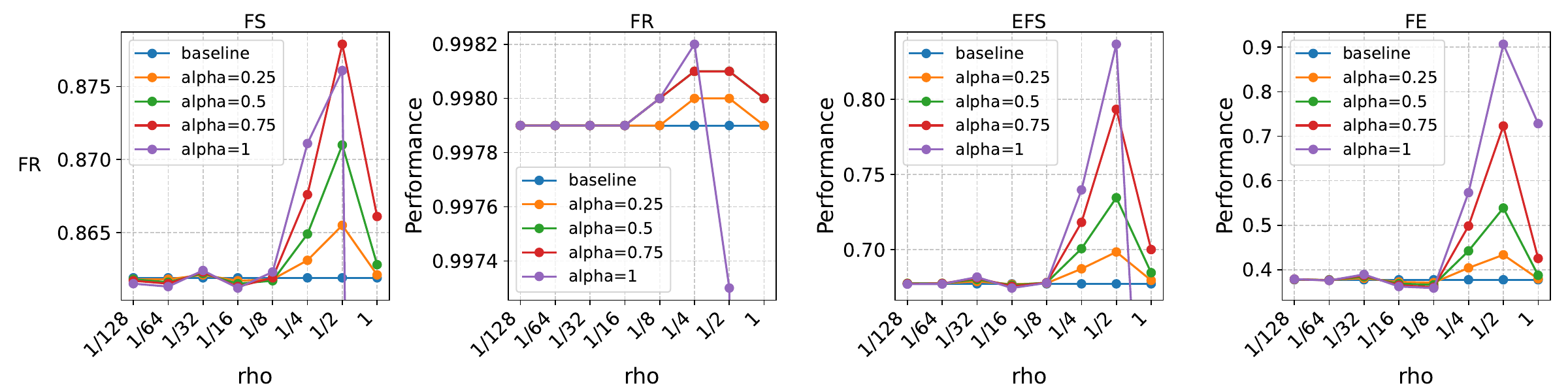}
  \caption{NAE hyperparameter sensitivity. Cross‑domain AUC (trained on FR, tested on EFS) varies with suppression ratio \(\rho\) and strength \(\alpha\). Optimal performance lies at \(\rho \in [1/4, 1/2]\) and \(\alpha=1.0\).}
  \label{fig:NAE-curve-FR} 
\end{figure*}

\subsubsection{Impact of NSP Hyperparameters}
\label{sec:ablation_nsp}

The subspace rank \(r\) and suppression strength \(\alpha\) jointly control the trade‑off between removing method‑specific shortcuts and preserving generalizable cues. The rank \(r\) determines the dimensionality of the suppressed subspace; \(\alpha\) scales the gradient projection along those directions.

\begin{figure}[h]
  \centering 
  \includegraphics[width=0.45\textwidth]{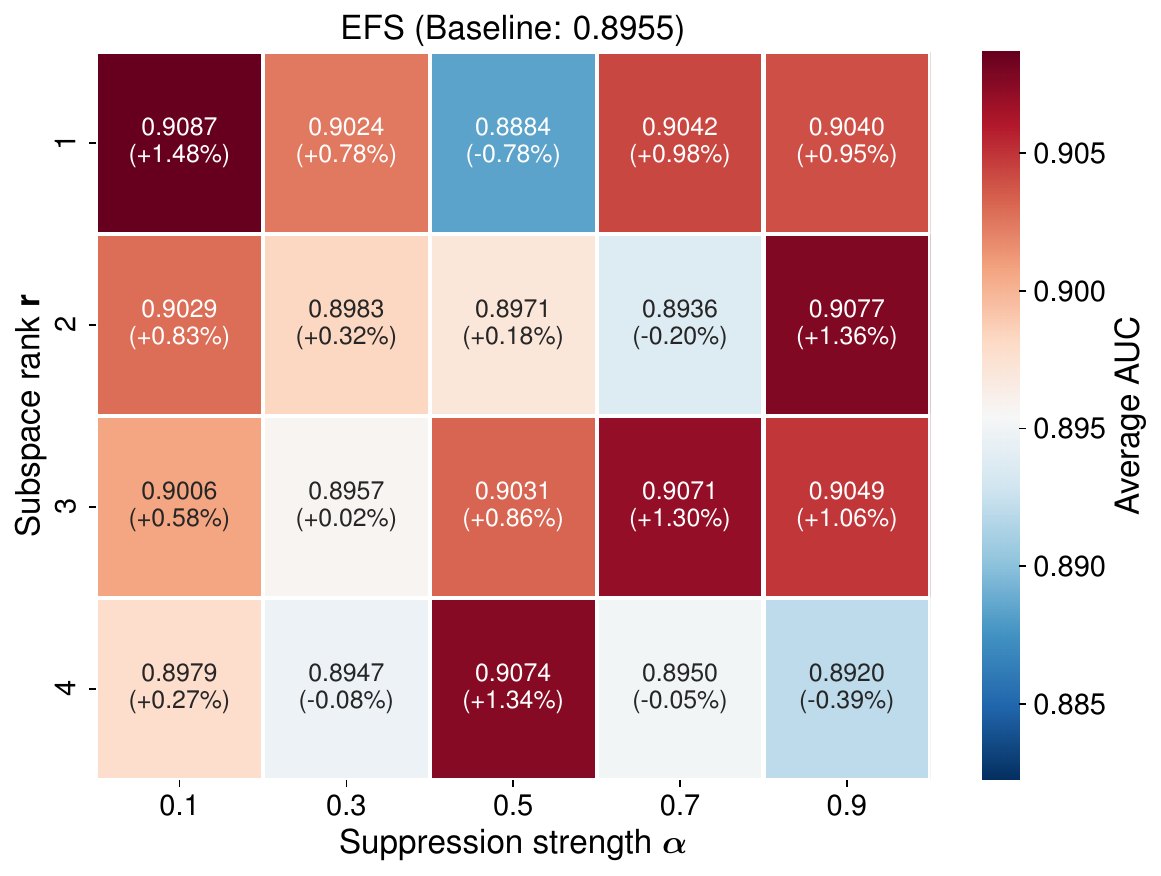}
  \caption{Impact of NSP hyperparameters. Cross‑domain AUC (trained on FS, tested on EFS) varies with subspace rank \(r\) and suppression strength \(\alpha\).}
  \label{fig:NSP-heatmap} 
  \vspace{-0.5cm}
\end{figure}

\textbf{Effect of rank \(r\).} Cross‑domain performance is sensitive to the choice of \(r\) (Figure~\ref{fig:NSP-heatmap}). Values in the range \(r=2\) or \(3\) yield the highest accuracy; decreasing \(r\) to \(1\) leaves insufficient suppression of method‑specific signals, while increasing \(r\) to \(4\) or higher risks suppressing generalizable forgery cues.

\textbf{Effect of suppression strength \(\alpha\).} Performance also exhibits clear sensitivity to \(\alpha\). Even a modest value like \(\alpha=0.1\) improves over the baseline, indicating that the suppression mechanism meaningfully influences gradient dynamics. Performance remains robust across a broad range (\(\alpha=0.1\) to \(0.9\)), with extreme values causing noticeable degradation.

Optimal configurations lie around \(r=3\) and \(\alpha=0.7\). The sensitivity to these choices reflects training‑time dynamics, where gradient projection amplifies even small changes in subspace alignment.

\subsubsection{Impact of NAE Hyperparameters}
\label{sec:ablation_NAE}

NAE introduces two hyperparameters: the suppression ratio \(\rho\) (proportion of neurons suppressed) and the suppression strength \(\alpha\) (scaling factor applied to selected neuron activations). We fix the method‑sensitive subspace rank to \(r=1\) and the localization layer to the last pointwise convolution; Appendix~\ref{app:arch_details} validates these choices.

\textbf{Effect of suppression ratio \(\rho\) and strength \(\alpha\).} The two parameters interact (Figure~\ref{fig:NAE-curve-FR}). Optimal performance occurs at \(\rho \in [1/4, 1/2]\) and \(\alpha=1.0\). Larger \(\rho\) (e.g., \(1\)) expands suppression to neurons carrying generalizable cues, hurting performance. Smaller \(\rho\) (e.g., \(1/16\)) removes too few method‑sensitive signals, leaving the model vulnerable to shortcuts. Complete removal (\(\alpha=1.0\)) works best; lower values (\(\alpha=0.25\)) weaken the intervention and reduce gains. Overall, NAE remains stable within this well‑defined region, and deviations in either direction degrade performance, underscoring the need for joint calibration.

\begin{figure}[t]
  \centering
  \includegraphics[width=0.4\textwidth]{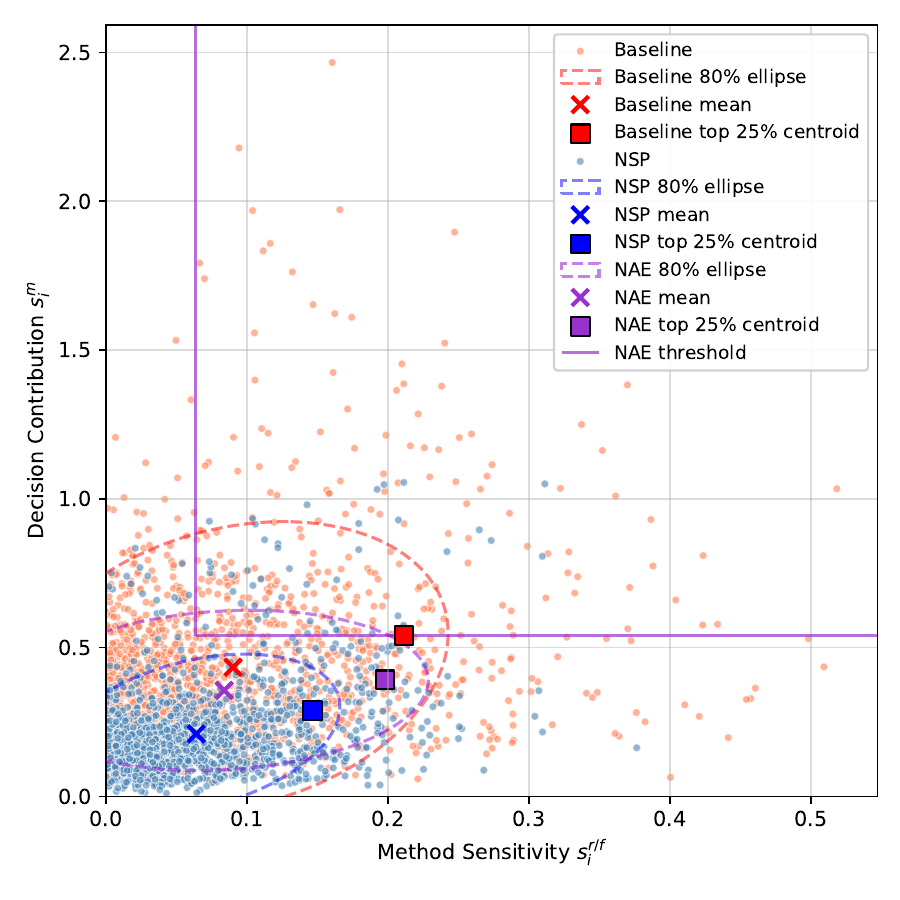}
    \setlength{\belowcaptionskip}{-0.5cm}
  \caption{Neuron-level analysis: decision contribution \(s_i^{\mathrm{rf}}\) vs. method sensitivity \(s_i^{\mathrm{m}}\). Confidence ellipses (one standard deviation) summarize each model's distribution. NSP reduces method sensitivity while preserving decision contribution; NAE directly eliminates the most offending neurons.}
  \label{fig:neuron_analysis}
\end{figure}

\subsection{Mechanism Analysis}

The generalization gains from NSP and NAE stem from a single cause: suppressing method‑specific shortcuts. Three lines of evidence support this claim.

\subsubsection{Neuron-Level Decoupling}

We analyze individual neurons in the final pointwise convolution layer, computing for each its decision contribution \(s_i^{\mathrm{rf}}\) and method sensitivity \(s_i^{\mathrm{m}}\). Figure~\ref{fig:neuron_analysis} visualizes the distribution.

NSP shifts the distribution downward along the method sensitivity axis while preserving decision contribution. The mean and top-25\% mean of \(s_i^{\mathrm{m}}\) decrease noticeably, and the confidence ellipse contracts vertically—a structural reshaping that suppresses method-specific shortcuts without severely sacrificing discriminative power.

NAE, by contrast, directly targets the most offending neurons. It suppresses the top \(\rho\) neurons with highest method sensitivity among those with above-median decision contribution (the neurons in the upper right corner of the threshold line are suppressed). This post-hoc correction removes extreme outliers rather than reshaping the overall distribution.

\subsubsection{Features Cluster by Forgery Method—Until Suppressed}

\begin{figure}[t]
  \centering 
  \subfloat[Xception]{\includegraphics[width=0.16\textwidth]{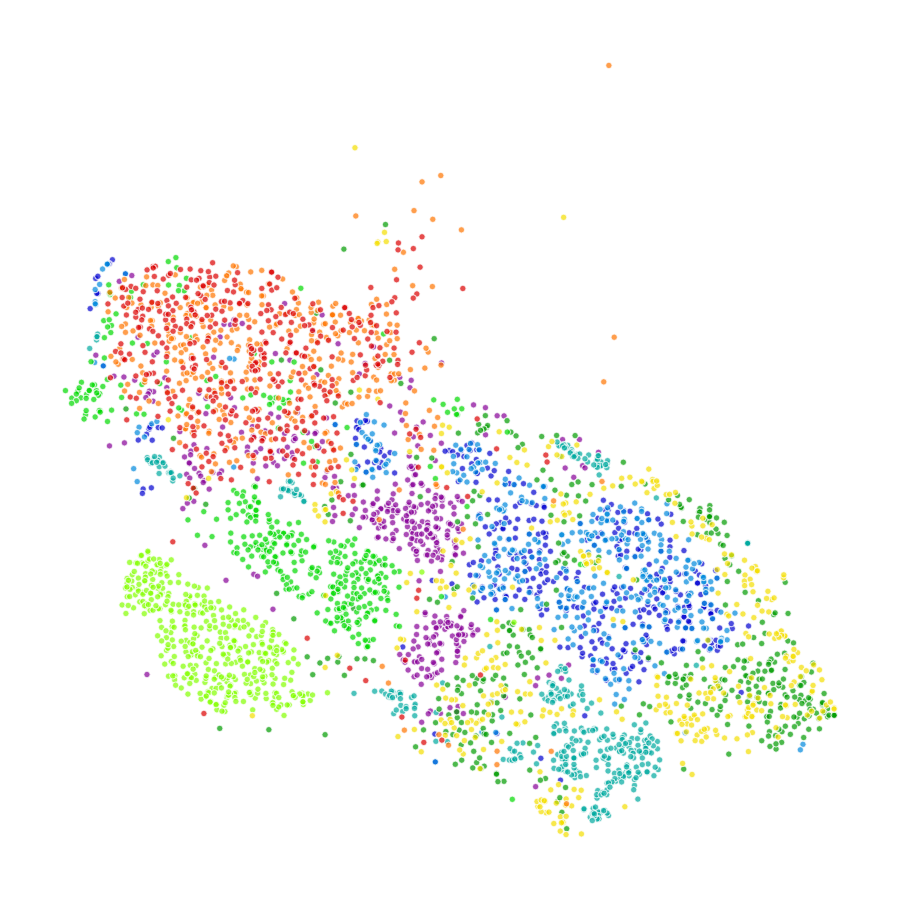}}
  \subfloat[$S^3$-NSP]{\includegraphics[width=0.16\textwidth]{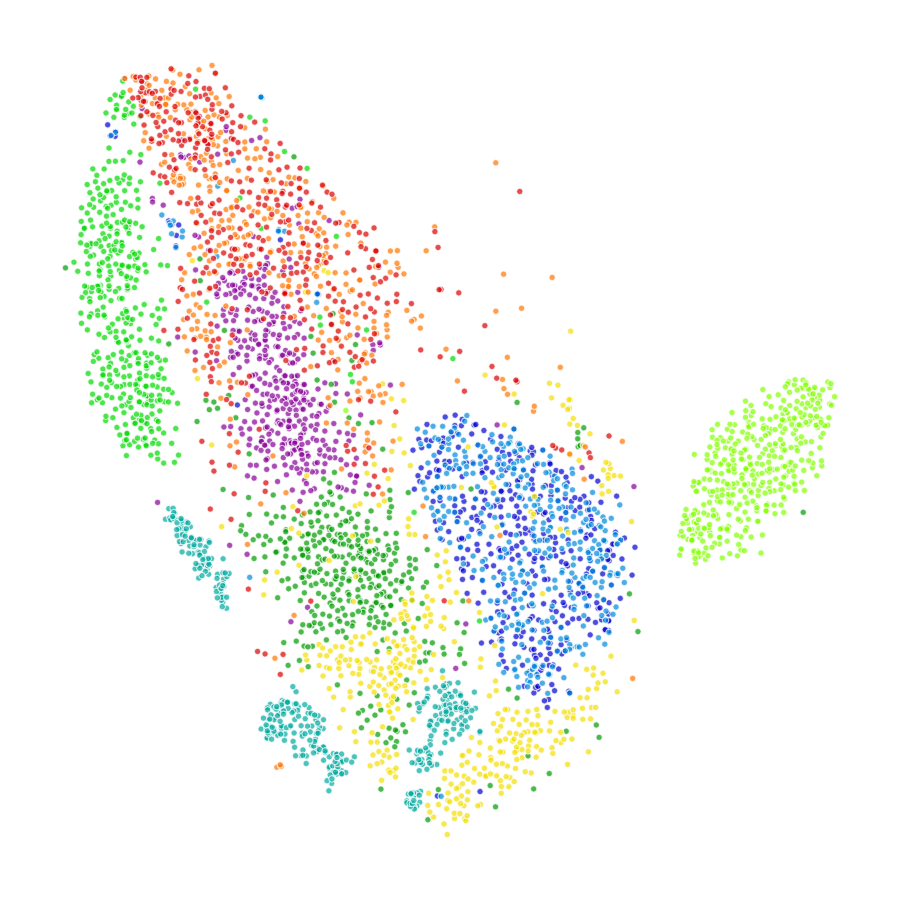}}
  \subfloat[$S^3$-NAE]{\includegraphics[width=0.16\textwidth]{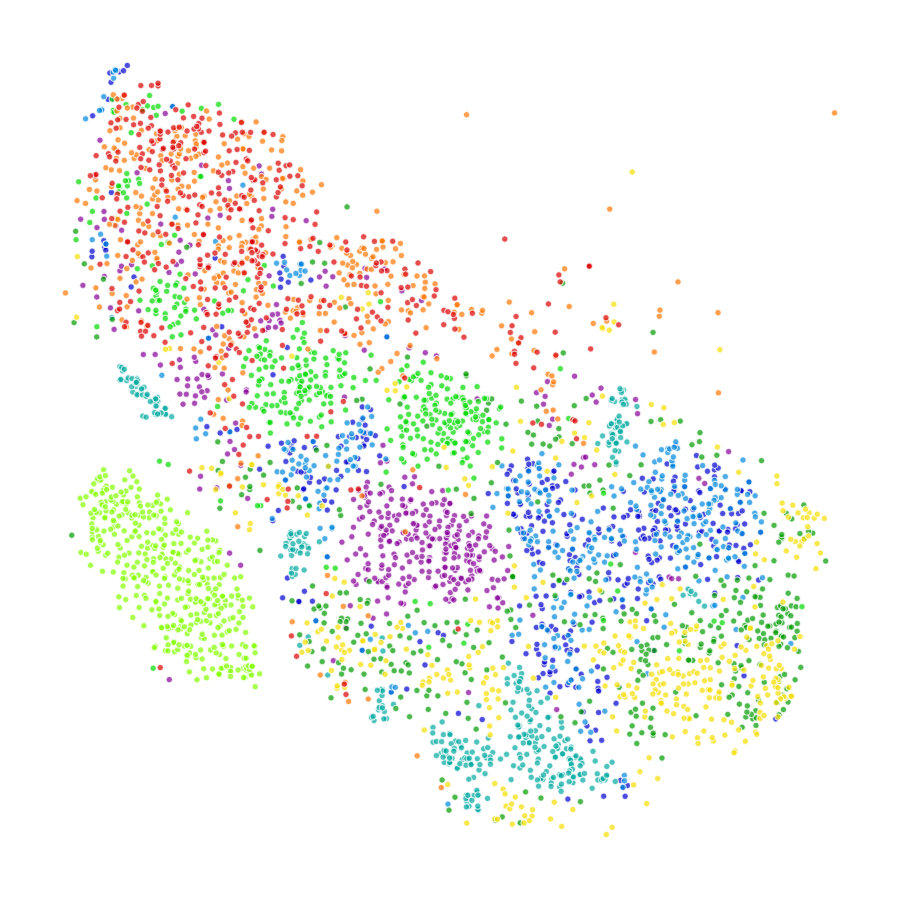}}

  \subfloat{\includegraphics[width=0.45\textwidth]{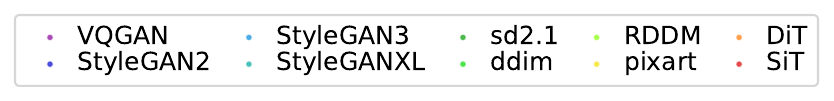}}
    \setlength{\belowcaptionskip}{-0.5cm}
    \caption{Features from an EFS‑trained model, colored by forgery method. Baseline (a) forms distinct method clusters; NSP (b) reconstructed feature distribution; NAE (c) substantially weakens them, though residual grouping remains.}
  \label{fig:tnse-EFS} 
\end{figure}

When trained on EFS, the baseline Xception organizes its feature space by manipulation method, despite having no access to method labels during training (Figure~\ref{fig:tnse-EFS}, a). This confirms that the model learns method‑specific artifacts rather than generalizable forensic cues. Applying NSP disrupts this structure: method clusters become less distinct, though partial separability remains (Figure~\ref{fig:tnse-EFS}, b). NAE weakens the clustering further, blurring method boundaries while leaving residual grouping (Figure~\ref{fig:tnse-EFS}, c). Together, these results show that both strategies suppress method‑specific features, with NAE acting as a stronger post‑hoc intervention.

\subsubsection{Shortcut Suppression Enables Cross‑Domain Transfer}

If method specific features are indeed non‑transferable, then suppressing them should improve real‑fake separation on unseen forgeries. This is exactly what we observe.

An Xception trained on FR and evaluated on EFS fails to separate real from fake in the baseline (Figure~\ref{fig:tnse-FR-EFS}, a). The model simply does not know what to look for. NSP produces a cleaner, better‑defined decision boundary (Figure~\ref{fig:tnse-FR-EFS}, b): suppressing method‑specific features during training forces the model to capture transferable forgery cues. NAE also improves over the baseline, yielding a more structured distribution (Figure~\ref{fig:tnse-FR-EFS}, c), though the separation is less pronounced—consistent with its inference‑time operation.

Together, these neuron-level changes explain the t-SNE observations: both methods decouple decision-critical features from method-specific artifacts, but through distinct mechanisms—NSP via learning dynamics that discourage shortcut formation, and NAE via direct attenuation at inference.

%% file: Deepfake_Sections/2-related-works.tex
\section{Related Works}

\subsection{Generalizable Deepfake Detection}

Early deepfake detectors identified forgery via low-level artifacts: frequency anomalies~\cite{qian2020thinking}, blending boundaries~\cite{durall2020watch,li2020face}, warping traces~\cite{li2019exposing}, or texture irregularities~\cite{liu2021spatial,wang2021representative}. These artifacts inherent to specific generation pipelines serve as discriminative cues. However, on unseen forgeries where such artifacts change or disappear, performance collapses.

To move beyond fixed artifacts, later efforts expanded training diversity through data synthesis~\cite{shiohara2022detecting,yan2024transcending,Hasanaath2024FSBIDD} or learned invariant representations via reconstruction~\cite{yan2023ucf}, frequency constraints~\cite{tan2024frequency}, identity disentanglement~\cite{dong2023implicit}, temporal consistency~\cite{haliassos2021lips,sun2021improving,zheng2021exploring}, and self-supervision~\cite{ni2022core,zhuang2022uia,zhao2021learning,larue2023seeable,qiao2024fully}. More recently, vision foundation models (VFMs) like CLIP have been adapted~\cite{smeu2025declip,yermakov2025unlocking,cozzolino2024raising}, leveraging their rich semantic priors to improve generalization.

Despite the diversity of these approaches, they share a common thread: they implicitly assume that generalization will emerge from more data, better invariance, or stronger priors, without ever identifying \emph{which} feature dimensions actually cause the failure. As a result, the model may still rely on spurious method-specific shortcuts that remain embedded in its representations. Our work takes a different stance: we explicitly localize the shortcut subspace via a lightweight probe and suppress it—either during training or at inference—rather than hoping it will not be learned.

\subsection{Disentanglement and Subspace Methods}

A more targeted line of work attempts to separate forensic cues from confounding factors by learning disentangled representations. UCF~\cite{yan2023ucf} decomposes features into forgery‑irrelevant, method‑specific, and common forgery components, using only the common part for detection. CADDM~\cite{dong2023implicit} targets identity leakage as a confounder. Effort~\cite{yan2024orthogonal} applies SVD to pre‑trained weights and updates only the residual subspace, preserving pre‑trained knowledge while learning forgery patterns. Low‑rank adaptation (LoRA)~\cite{hu2021lora} injects trainable low‑rank matrices into frozen backbones.

The common philosophy here is to \emph{learn around} shortcut information—by isolating a “safe” subspace or confining updates to a low‑rank manifold. However, these methods do not explicitly remove the shortcut components from the representation; the shortcuts remain present and can still influence the final decision. The question of \emph{which} directions in the feature space encode non‑transferable shortcuts—and how to suppress them directly—remains unanswered. In contrast, our framework (i) explicitly constructs the shortcut subspace via a lightweight probe, (ii) projects it out during training, and (iii) provides a training‑free inference‑time suppression that also reveals which neurons encode the shortcuts.

\begin{figure}[t]
  \centering 
  \subfloat[Xception]{\includegraphics[height=0.16\textwidth]{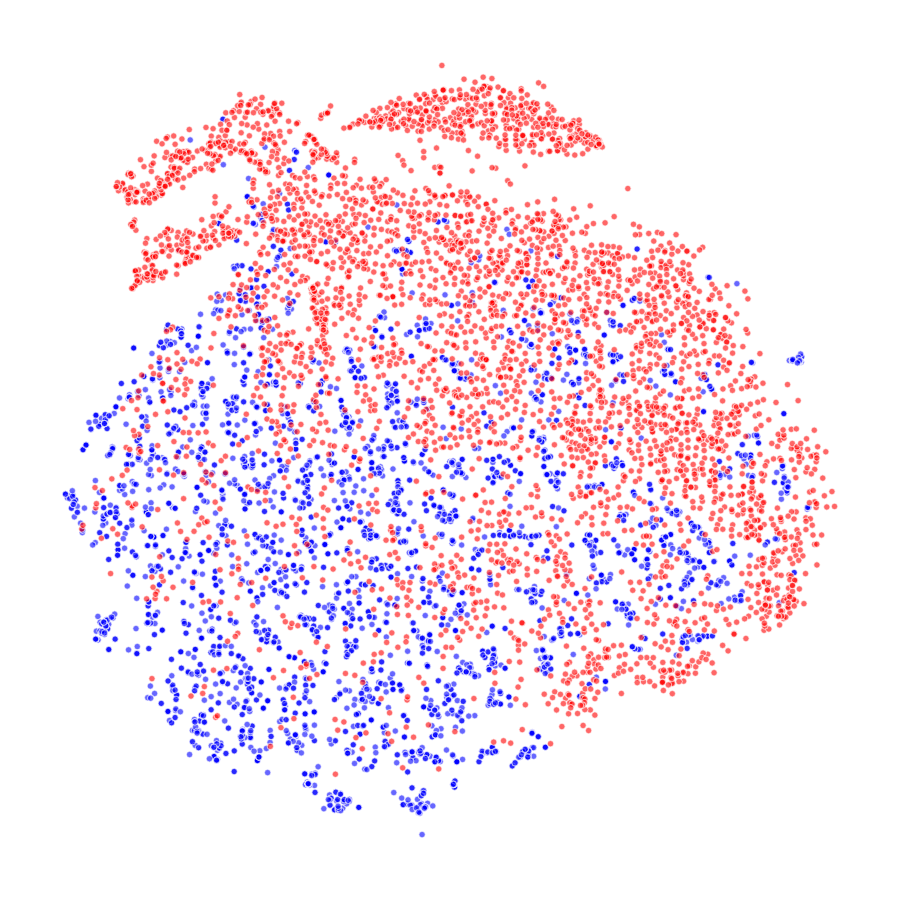}}
  \subfloat[$S^3$-NSP]{\includegraphics[height=0.16\textwidth]{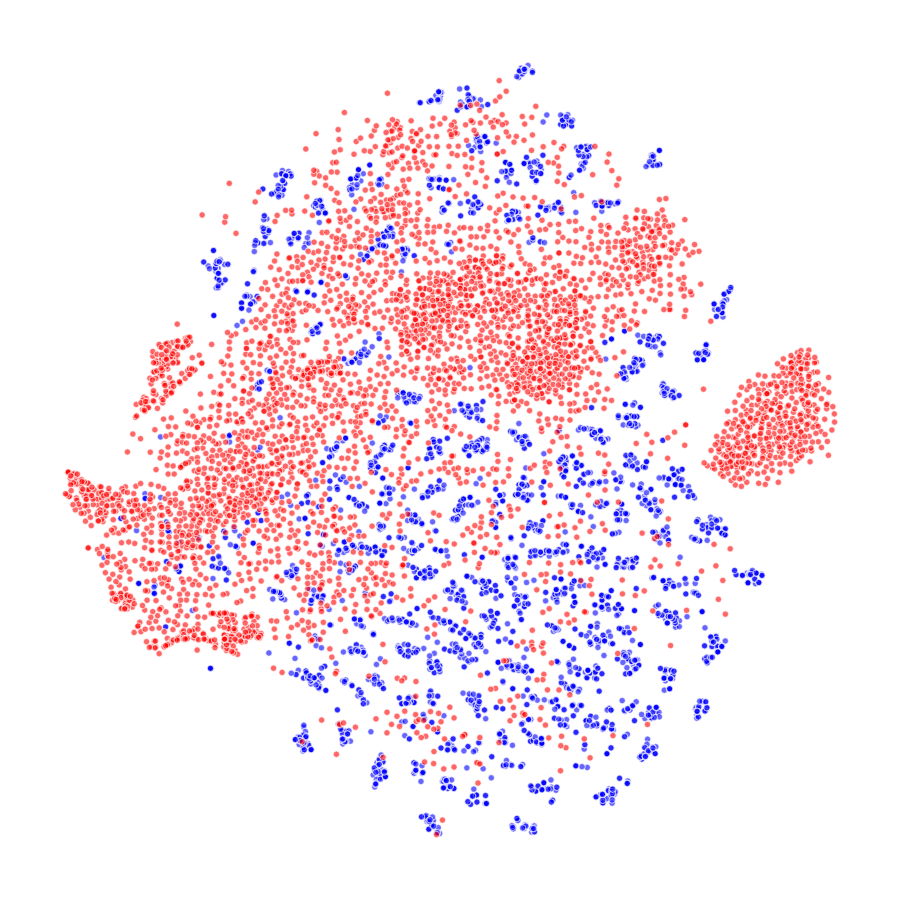}}
  \subfloat[$S^3$-NAE]{\includegraphics[height=0.16\textwidth]{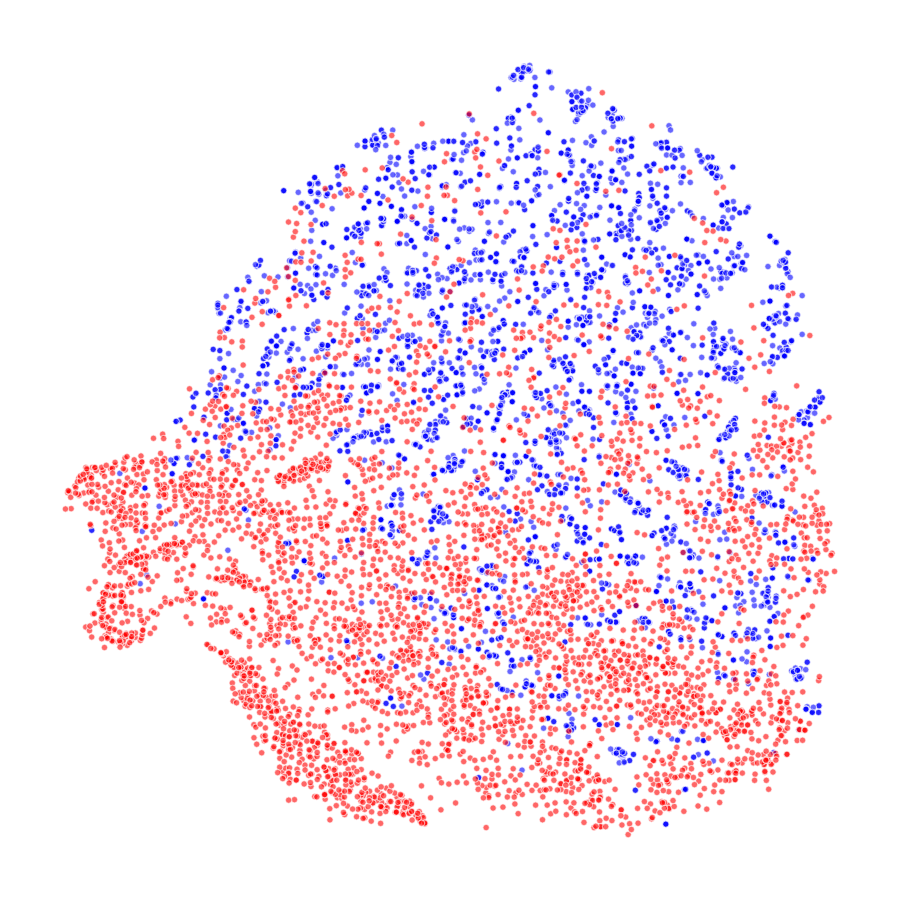}}

    \subfloat{\includegraphics[width=0.15\textwidth]{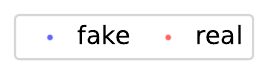}}
      \setlength{\belowcaptionskip}{-0.5cm}
  \caption{Features from an FR‑trained model tested on EFS, colored by real/fake. Baseline (a) shows no separation; NSP (b) establishes a clear boundary; NAE (c) improves but less than NSP.}
  \label{fig:tnse-FR-EFS} 
\end{figure}

%% file: Deepfake_Sections/5-conclusion.tex
\section{Conclusion}

In this work, we identify and suppress a key cause of poor cross-method generalization in deepfake detection: \emph{method-specific shortcuts}. Our core insight is that transferable forensic cues and method-specific shortcuts occupy complementary subspaces. To address this, we propose the \textbf{Shortcut Subspace Suppression ($S^3$) framework}, which explicitly characterizes these shortcuts as a low-dimensional subspace. The framework provides two complementary strategies: Nullspace Projection (NSP) suppresses shortcut-aligned gradients during training, steering the model toward generalizable features; Neuron Activation Editing (NAE) attenuates shortcut-sensitive neurons at inference without retraining, offering a plug-and-play boost. Extensive experiments on DF40 demonstrate consistent gains across backbones and forgery types with negligible overhead. Explicitly suppressing the shortcut subspace forces detectors to rely on transferable evidence, opening new directions for deepfake detection and broader out-of-distribution generalization tasks.

%% file: Deepfake_Sections/6-Appendix.tex
\section{Algorithm Overview}
\label{app:algorithm}

This appendix presents the complete procedure of our proposed \shortname~framework, as summarized in Algorithm~\ref{alg:s3}.

\begin{algorithm}[h]
\caption{\shortname: A Unified Framework for Shortcut Subspace Suppression}
\label{alg:s3}
\KwIn{Training set $\mathcal{D}_{\mathrm{train}}$, test sample $x$, feature extractor $\phi_\theta$, classifier $\psi$, subspace rank $r$, NSP strength $\alpha_{\mathrm{nsp}}$, NAE strength $\alpha_{\mathrm{nae}}$, neuron ratio $\rho$}
\KwOut{Prediction $\hat{y}$}

\BlankLine
\textbf{Strategy I: Training-time Subspace Suppression (NSP)} \\
\For{each training iteration}{
    \uIf{epoch $\ge T_1$}{
        Train linear probe $W_m$  \\
        Perform SVD on $W_m$ (Eq.~\ref{eq:svd}) to update $U_r$ and projection $\mathbf{P}=U_rU_r^\top$
    }
    Extract features, compute detection loss, and obtain gradient $\nabla\mathcal{L}$ \\
    Apply gradient projection (Eq.~\ref{eq:nsp_proj}) \\
    Update model parameters
}

\BlankLine
\textbf{Strategy II: Inference-time Neuron Suppression (NAE)} \\
Train linear probe $W_m$ \\
Perform SVD on $W_m$ to obtain $U_r$ \\
Select localization layer and compute neuron scores $s_i^{\text{r/f}}$, $s_i^{\text{m}}$ (Eq.~\ref{eq:srf}, \ref{eq:sm}) \\
Identify shortcut neuron set $\mathcal{I}$ \\
Extract activation tensor $\mathcal{A}$ from the localization layer \\
\For{each $i \in \mathcal{I}$}{ $\mathcal{A}[:, i] \leftarrow \mathcal{A}[:, i] \cdot (1 - \alpha_{\mathrm{nae}})$} 
\BlankLine
Forward edited activations to obtain prediction $\hat{y}$
\end{algorithm}

\section{Additional Details of NSP}
\label{app:nsp_details}

\subsection{Theoretical Analysis}

A natural concern is whether suppressing gradient components along $\operatorname{span}(\mathbf{U}_r)$ harms the primary real/fake discrimination. We argue that it does not. Consider a first‑order Taylor expansion of the detection loss $\mathcal{L}$ around the current feature representation $\mathcal{Z}$:

\[
\mathcal{L}(\mathcal{Z} - \eta \nabla_{\mathcal{Z}} \mathcal{L}_{\text{new}}) \approx \mathcal{L}(\mathcal{Z}) - \eta \langle \nabla_{\mathcal{Z}} \mathcal{L}, \nabla_{\mathcal{Z}} \mathcal{L}_{\text{new}} \rangle,
\]

where $\eta$ is the learning rate. Substituting 

\[
\nabla_{\mathcal{Z}} \mathcal{L}_{\text{new}} = \nabla_{\mathcal{Z}} \mathcal{L} - \alpha \mathbf{P} \nabla_{\mathcal{Z}} \mathcal{L}
\]

gives

\[
\langle \nabla_{\mathcal{Z}} \mathcal{L}, \nabla_{\mathcal{Z}} \mathcal{L}_{\text{new}} \rangle = \|\nabla_{\mathcal{Z}} \mathcal{L}\|_F^2 - \alpha \|\mathbf{U}_r^\top \nabla_{\mathcal{Z}} \mathcal{L}\|_F^2.
\]

Both terms are non‑negative, so the inner product is non‑negative. Thus, for sufficiently small $\eta$, the loss does not increase—the projection does not counteract the utility objective. The model can still learn from the residual component $\mathbf{P}^\perp \nabla_{\mathcal{Z}} \mathcal{L}$, which captures gradients orthogonal to the method‑sensitive subspace. This derivation also implies that $\alpha$ must satisfy $\alpha \leq 1$ to ensure the inner product remains non‑negative; larger values would risk reversing the gradient component along the shortcut subspace, potentially harming detection. Empirically, as shown in Sec.~\ref{sec:ablation_nsp}, in‑domain performance remains virtually unchanged with $\alpha \in [0,1]$, confirming that the suppressed directions are indeed shortcuts rather than generalizable forensic cues.

\subsection{Training Schedule}
\label{app:nsp_training}

To ensure accurate subspace construction while allowing the model to benefit from domain‑sensitive cues in early training, we adopt a staged training schedule.

\textbf{Stage I: Main Task Warm‑up.}
In the first $T_1$ epochs, only the main real/fake classifier is trained. The method probe is frozen, and NSP is disabled. This stage allows the backbone to learn basic discriminative features, leveraging method‑specific signals that help the model converge.

\textbf{Stage II: Alternating Training for Probe Learning.}
After the warm‑up, we enable the alternating freezing strategy described in Sec.~\ref{sec:nsp}. In each training iteration, we alternate between updating the real/fake classifier (with the probe frozen) and updating the method probe (with the real/fake head and backbone frozen). This alternating schedule ensures that the probe learns method‑specific signals without interfering with the primary detection task, while keeping the probe weights informative for subspace construction. The subspace $U_r$ is recomputed periodically after every probe training epoch to reflect the evolving probe weights.

\textbf{Stage III: Progressive NSP Activation.}
Once the alternating training is stable, we activate NSP while continuing the alternating schedule. To avoid injecting a noisy subspace early, we employ a warm‑up strategy for the suppression strength: $\alpha$ is set to $0$ initially and linearly increased to its target value over $T_3$ epochs. This gradual ramp‑up ensures that gradient suppression does not disrupt learning while gradually steering the model away from method‑specific shortcuts.

In our experiments, we set $T_1 = 3$ epochs, $T_2 = 2$ alternating updates, and $T_3 = 5$ epochs. These values can be adjusted based on dataset size and convergence behavior.

\section{Detailed Architecture Adaptation for NAE}
\subsection{Detailed Derivation of the Linear Approximation in NAE}
\label{app:NAE_derivation}

Let $\mathbf{z} \in \mathbb{R}^d$ be the feature representation before the classification layer, and let $\mathbf{n}_{\text{r/f}} = \mathbf{W}_1 - \mathbf{W}_0$ be the normal vector of the decision boundary, where $\mathbf{W}_0, \mathbf{W}_1 \in \mathbb{R}^d$ are the binary classification head weights. The model's confidence in classifying a sample as fake is proportional to $\mathbf{n}_{\text{r/f}}^\top \mathbf{z}$.

In the backbone, $\mathbf{z}$ derives from the activations $\mathbf{a} \in \mathbb{R}^h$ of an earlier layer through a sequence of transformations. For a localization layer where the mapping from $\mathbf{a}$ to $\mathbf{z}$ is approximately linear, we write:

\[
\mathbf{z} = \mathbf{W}_{\text{loc}} \mathbf{a} + \mathbf{b},
\]

where $\mathbf{W}_{\text{loc}} \in \mathbb{R}^{d \times h}$ is the weight matrix and $\mathbf{b} \in \mathbb{R}^d$ is the bias. Each column $\mathbf{w}_i \in \mathbb{R}^d$ corresponds to the weight vector of neuron $\mathbf{a}_i$. The bias contributes a constant offset across samples and does not affect the relative influence of different input neurons on the decision. Substituting into the decision confidence:

\[
\mathbf{n}_{\text{r/f}}^\top \mathbf{z} = \mathbf{n}_{\text{r/f}}^\top (\mathbf{W}_{\text{loc}} \mathbf{a} + \mathbf{b}) = (\mathbf{W}_{\text{loc}}^\top \mathbf{n}_{\text{r/f}})^\top \mathbf{a} + \mathbf{n}_{\text{r/f}}^\top \mathbf{b}.
\]

Define $\mathbf{c} = \mathbf{W}_{\text{loc}}^\top \mathbf{n}_{\text{r/f}} \in \mathbb{R}^m$ and $\text{const} = \mathbf{n}_{\text{r/f}}^\top \mathbf{b}$. Then:

\[
\mathbf{n}_{\text{r/f}}^\top \mathbf{z} = \mathbf{c}^\top \mathbf{a} + \text{const}.
\]

Thus $\mathbf{c}_i$ approximates the contribution coefficient of neuron $\mathbf{a}_i$ to the final logit difference. Adjusting $\mathbf{a}_i$ proportionally changes the decision value, justifying the neuron suppression strategy.

\subsection{Principles for Localization Layer Selection}
\label{app:principles}

The effectiveness of Neuron Activation Editing (NAE) relies on choosing an appropriate localization layer—the layer whose activations we directly attenuate. We follow three guiding principles.

\paragraph{Dimensionality Matching.}
To compute the decision contribution \(s_i^{\text{r/f}}\) and method sensitivity \(s_i^{\text{m}}\) for a neuron, its weight vector \(\mathbf{w}_i\) must reside in the same space as the decision normal vector \(\mathbf{n}_{\text{r/f}}\) and the method-sensitive subspace \(\mathbf{U}_r\). Concretely, if the feature representation before the classification head has dimension \(d\), the localization layer must output a tensor that is either directly the feature vector \(\mathbf{z} \in \mathbb{R}^d\) or can be reshaped/aggregated into that dimension. Otherwise, the inner products are ill-defined, making neuron-level analysis impossible. Therefore, we restrict our choice to layers whose output dimension matches the input dimension of the classifier head. 

\paragraph{Applicability of the Linear Assumption.}
Our theoretical derivation assumes a strictly linear transformation \(\mathbf{z} = \mathbf{W}_{\text{loc}} \mathbf{a}\). The closer the layer is to the classification head, the fewer subsequent non-linear transformations intervene, and thus the more accurate the linear approximation. Shallow layers, in contrast, require \(\mathbf{a}\) to pass through many non-linear operations before reaching the head, weakening both the validity of the linear assumption and the interpretability of suppressing \(\mathbf{a}\) directly.

\paragraph{Consideration of Residual Connections.}
In networks with residual connections, one should select a node that all subsequent computations must pass through. For example, in residual networks, the activation between two residual blocks is such a bottleneck; if a branch inside a residual block is chosen, part of the information may bypass the intervention through other paths. Therefore, we prioritize linear layers where the intervened neurons lie on a path without bypass branches that all subsequent calculations depend on.

Based on these principles, we select the last linear transformation layer in the backbone that is directly associated with the classification feature dimension as the localization layer. Empirical results show that moving the localization layer forward significantly reduces the generalization improvement of NAE, validating the effectiveness of this choice.

\subsection{Architecture-Specific Localization Layers}
\label{app:arch_details}

Following the principles outlined above, we provide the specific localization layer choices for four typical backbone architectures used in NAE.

\begin{itemize}
    \item \textbf{Xception}  
    After multiple SeparableConv blocks, the feature tensor passes through a pointwise convolution layer before global average pooling. This layer is a \(1\times1\) convolution, with input shape \((B, 1536, H, W)\) and output shape \((B, 2048, H, W)\). The weight matrix has shape \((2048, 1536)\); we take the column vectors \(\mathbf{w}_i \in \mathbb{R}^{2048}\) as the weights for each input neuron (1536 in total). The input feature dimension for the method probe is \(d = 2048\), so \(\mathbf{U}_r \in \mathbb{R}^{2048 \times r}\) can be directly multiplied with \(\mathbf{w}_i\).

    \item \textbf{ResNet50}  
    In the last Bottleneck, the downsampling convolutional layer expands the input channels from 1024 to 2048 while changing the spatial size. This layer is a \(1\times1\) convolution, with input shape \((B, 1024, H, W)\) and output shape \((B, 2048, H/2, W/2)\). The weight matrix has shape \((2048, 1024)\); the column vectors \(\mathbf{w}_i \in \mathbb{R}^{2048}\) correspond to the weights of the 1024 input neurons. Here \(d = 2048\), matching the probe’s input dimension. However, this choice means that we can only intervene on the skip branch, and the effect is not as significant as that on Xception, which has also been confirmed in our experiments (see Section~\ref{sec:generalizable_exp}).

    \item \textbf{EfficientNet-B4}  
    The convolutional layer before global average pooling (conv\_head) is a \(1\times1\) convolution, with input shape \((B, 448, H, W)\) and output shape \((B, 1792, H, W)\). The weight matrix has shape \((1792, 448)\); the column vectors \(\mathbf{w}_i \in \mathbb{R}^{1792}\) correspond to the weights of the 448 input neurons. Here \(d = 1792\), matching the probe’s input dimension.

    \item \textbf{CLIP (ViT)}  
    In the Vision Transformer, no single deep layer places the intervened neurons at a clear information bottleneck; therefore we select an alternative location. The MLP module of the last encoder layer consists of two fully connected layers: fc1 expands the feature dimension from 768 to 3072, and fc2 compresses it back to 768. We choose fc2 as the localization layer; its weight matrix has shape \((768, 3072)\), and the input neurons correspond to the 3072 neurons in the MLP intermediate layer. The output dimension \(d = 768\) matches the classification feature dimension. We take the column vectors \(\mathbf{w}_i \in \mathbb{R}^{768}\) as the weights for each MLP intermediate neuron, and the method probe’s \(\mathbf{U}_r \in \mathbb{R}^{768 \times r}\) can be directly aligned with them. Suppressing these neurons effectively weakens method-sensitive intermediate features in the MLP, thereby indirectly affecting the final classification.
\end{itemize}

\subsection{Empirical Validation of Localization Layer Selection}

To verify that the chosen localization layer (the final MLP layer in CLIP) yields the best cross-domain generalization, we conduct an ablation study across all 12 transformer blocks of CLIP (ViT-B/16). For each block, we treat its output features as the localization layer and apply NAE with the same hyperparameters (\(\rho=1/4, \alpha=1.0\)). All models are trained on FR(FF) and tested on FS, FR, EFS, and FE subsets of DF40.

\begin{figure}[ht]
\centering
\includegraphics[width=0.45\textwidth]{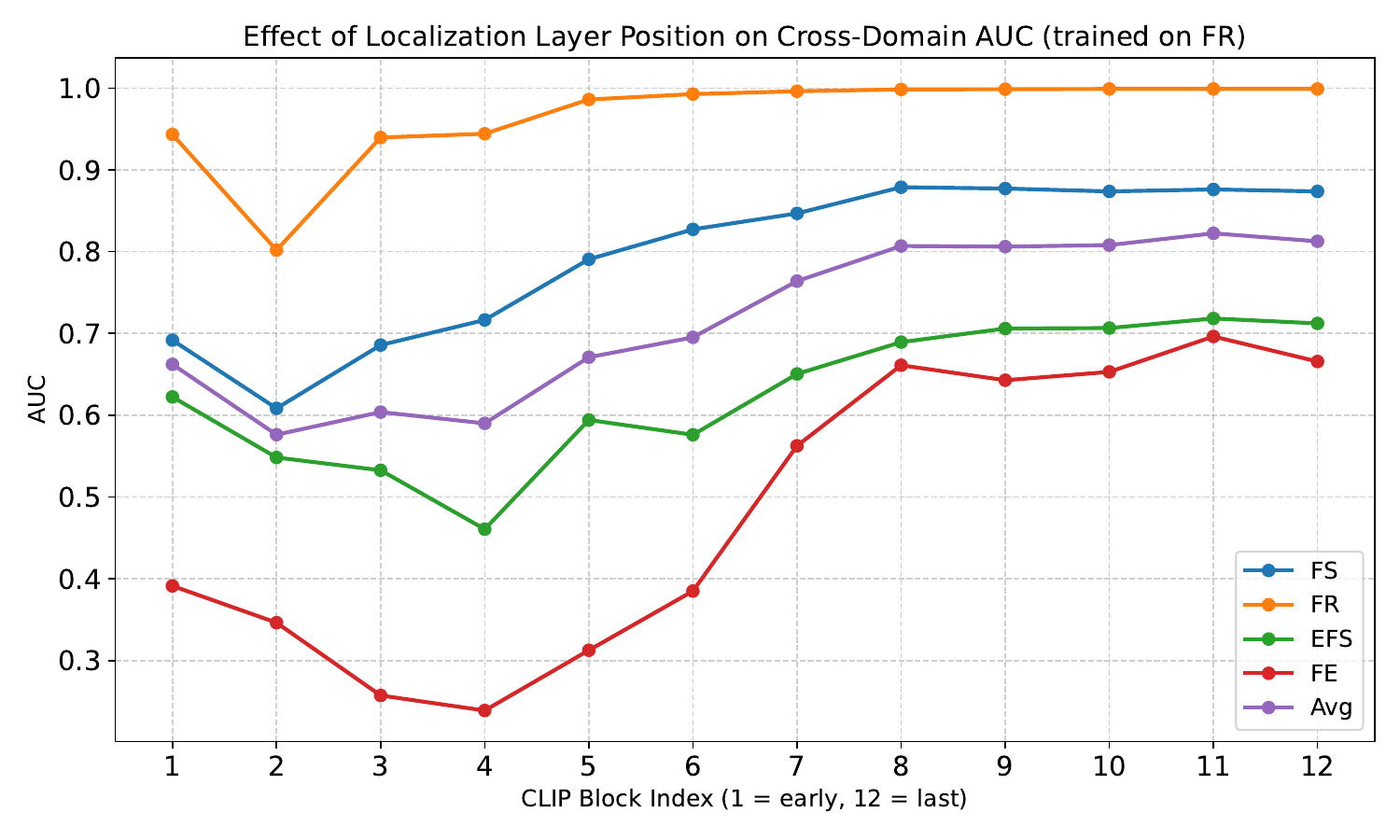}
\caption{Average AUC as a function of CLIP block index. The dashed line indicates the baseline (no suppression).}
\label{fig:clip_localization_trend}
\end{figure}

Figure~\ref{fig:clip_localization_trend} visualizes the average AUC for each block. As the block index increases (moving from early to late layers), cross-domain performance generally improves. Blocks 2–4 show the poorest generalization, especially on EFS and FE. Block 1 yields slightly higher AUC than Blocks 2–4. This pattern reflects a key insight: in shallow layers, the linearity assumption in Eq.~(8) does not hold—the weights of the real/fake classifier and method probe are not aligned with these early features. Consequently, the neurons selected for suppression are not genuinely decision-critical or method-sensitive; suppressing them introduces near-random perturbations that have limited impact on the final output. In Blocks 2–4, the features begin to partially align with the classification heads but remain misaligned enough that suppression causes unintended interference, degrading performance further. From Block 8 onward, the features become sufficiently aligned, allowing NAE to effectively remove shortcut neurons and improve generalization. The deepest blocks (11–12) achieve the highest and most stable AUC, confirming our three design principles mentioned before.

Based on these results, we fix the localization layer as the last MLP layer (fc2) of the final transformer block for CLIP, and as the last pointwise convolution layer for CNNs. The subspace rank is set to \(r=1\) for all architectures, as further ablation confirms that a single principal direction already captures the dominant method-sensitive subspace.

\section{Additional Information on Dataset}

\begin{table*}[ht]
\centering
\caption{Summary of the 32 deepfake methods from the DF40 dataset employed in this study}
\label{tab:fake_methods}
\tabcolsep=0.07cm
\renewcommand{\arraystretch}{0.6}
\tiny
\resizebox{0.92 \textwidth}{!}{ 
\begin{tabular}{c|c|ccccc}
\toprule[1.2pt]
\textbf{Type} & \textbf{ID} & \textbf{Method} & \textbf{Sub-Types} & \textbf{Venue} & \textbf{Real Data Source} & \textbf{Data Used} \\
\midrule
\multirow{9}{*}{Face-swapping (FS)} & 1  & FSGAN      & Parsing mask        & ArXiv 2019     & FF++ & Train \& Test \\
 & 2  & FaceSwap     & Graphic based      & None           & FF++ & Train \& Test \\
 & 3  & SimSwap      & Disentangle        & ICCV 2019      & FF++ & Train \& Test \\
 & 4  & InSwapper  & Used in Roop  & None           & FF++ & Train \& Test \\
 & 5  & BlendFace   & Disentangle        & ICCV 2023      & FF++ & Train \& Test \\
 & 6  & UniFace     & Disentangle        & ECCV 2022      & FF++ & Train \& Test \\
 & 7  & MobileSwap  & Lightweight        & AAAI 2022      & FF++ & Train \& Test \\
 & 8  & e4s         & Disentangle        & CVPR 2023      & FF++ & Train \& Test \\
 & 9  & FaceDancer  & Disentangle        & WACV 2023      & FF++ & Train \& Test \\
\midrule
\multirow{12}{*}{Face-reenactment (FR)} & 10 & FOMM        & Image Driven       & NeurIPS 2019   & FF++ & Train \& Test \\
 & 11 & FS\_vid2vid  & Landmark Driven    & ArXiv 2019     & FF++ & Train \& Test \\
 & 12 & Wav2Lip      & Audio Driven       & MM 2020        & FF++ & Train \& Test \\
 & 13 & MRAA         & Image Driven       & CVPR 2021      & FF++ & Train \& Test \\
 & 14 & OneShot     & Image Driven       & CVPR 2021      & FF++ & Train \& Test \\
 & 15 & PIRender     & Image Driven       & ICCV 2021      & FF++ & Train \& Test \\
 & 16 & TPSMM        & Image Driven       & CVPR 2022      & FF++ & Train \& Test \\
 & 17 & LIA         & Image Driven       & ICLR 2022      & FF++ & Train \& Test \\
 & 18 & DaGAN        & Image Driven       & CVPR 2022      & FF++ & Train \& Test \\
 & 19 & SadTalker    & Audio Driven       & CVPR 2023      & FF++ & Train \& Test \\
 & 20 & MCNet       & Image Driven       & ICCV 2023      & FF++ & Train \& Test \\
 & 21 & HyperReenact & Image Driven       & ICCV 2023      & FF++ & Train \& Test \\
\midrule
\multirow{10}{*}{Entire Face Synthesis (EFS)} & 22 & VQGAN        & GAN based          & CVPR 2021      & FF++ & Train \& Test \\
 & 23 & StyleGAN2    & GAN based          & ArXiv 2019     & FF++ & Train \& Test \\
 & 24 & StyleGAN3    & GAN based          & NeurIPS 2021   & FF++ & Train \& Test \\
 & 25 & StyleGAN-XL  & GAN based          & SIGGRAPH 2022  & FF++ & Train \& Test \\
 & 26 & SD-2.1       & Latent Diffusion   & CVPR 2022      & FF++ & Train \& Test \\
 & 27 & DDPM        & Latent Diffusion   & NeurIPS 2020   & FF++ & Train \& Test \\
 & 28 & RDDM        & Latent Diffusion   & ArXiv 2023     & FF++ & Train \& Test \\
 & 29 & PixArt-$\alpha$  & Latent Diffusion   & ICLR 2024      & FF++ & Train \& Test \\
 & 30 & DiT-XL/2     & Latent Diffusion   & ICCV 2023      & FF++ & Train \& Test \\
 & 31 & SiT-XL/2      & Latent Diffusion   & ArXiv 2024     & FF++ & Train \& Test \\
\midrule
Face editing (FE) & 32 & e4e          & StyleGAN based     & SIGGRAPH 2021  & FF++ & Test Only \\
\bottomrule[1.2pt]
\end{tabular}
}
\end{table*}

While the main text presents an overview of the datasets used, this appendix provides a more comprehensive description of the data processing and sampling strategies adopted to ensure a fair and rigorous evaluation.

\textbf{Clarification of Forgery Categories.}
Throughout this paper, we refer to FS (Face Swapping), FR (Face Reenactment), EFS (Entire Face Synthesis), and FE (Face Editing) as broad categories of forgery techniques. Each category contains multiple specific manipulation methods, as enumerated in Table~\ref{tab:fake_methods}. For example, FS includes FSGAN, FaceSwap, SimSwap, InSwapper, and others; FR includes FOMM, Wav2Lip, SadTalker, and others; EFS includes StyleGAN2, SD-2.1, DDPM, and others; FE includes e4e. This categorization follows the standard protocol of DF40. When the paper states “training on FS” or “testing on FR”, it means training or testing on all methods belonging to that category.

\textbf{Video‑Level Splitting.}
All experiments are conducted on the FF domain of DF40, which uses the original real videos from FaceForensics++ (FF++) as the source of authentic content. To prevent any information leakage between training and testing, the split is performed at the video level. Specifically, the official FF++ video split (720 videos for training, 140 for validation/testing) is used, and all forged samples are generated from the corresponding real videos in the respective split. Consequently, no real video—or any of its associated forged variants—appears in both the training and testing sets.

\textbf{Controlling per‑Video Sampling.}
A single video typically contains hundreds of frames, and consecutive frames often exhibit high visual similarity. Sampling too many frames from the same video would effectively present the model with highly correlated samples within a single epoch, which could bias the learning process. To mitigate this, we limit the number of frames sampled per video. Specifically, for each manipulation method, we randomly select a maximum of 10 frames per video. This constraint ensures that the training set covers a diverse set of video content without over‑representing any particular video.

\textbf{Balanced Sampling Across Forgery Methods.}
The DF40 dataset provides an imbalanced number of samples across different manipulation techniques. To prevent the model from learning a bias toward methods with larger data volumes, we adopt a balanced sampling strategy. For each training set (FS, FR, or EFS), we uniformly sample a fixed number of real–fake pairs from each forgery method. The sampling is performed across different videos to maximize content diversity. Specifically, we sample:

\begin{itemize}
\item \textbf{Training set:} 1,000 real–fake pairs per forgery method.
\item \textbf{Validation set:} 200 real–fake pairs per forgery method.
\item \textbf{Testing set:} 400 real–fake pairs per forgery method.
\end{itemize}

Each pair consists of a fake image and its corresponding source real image. This paired structure is essential for the reconstruction and contrastive losses used in our framework.

\textbf{Avoiding Shortcut Learning via Batch‑Level Pairing.}
When training with paired real–fake samples, a naive implementation could allow the model to rely on trivial correlations, such as remembering the identity or background from the real image and simply comparing it with the fake image. To prevent such shortcut learning, we ensure that for every real–fake pair used in the loss computation, the corresponding real sample is guaranteed to appear in the same training batch. However, we do not expose the model to the exact paired real image at the input level; instead, the pair is used internally for the reconstruction and contrastive objectives. This design forces the model to learn the actual forgery traces rather than exploiting simple identity correspondences.

These processing choices collectively ensure that our evaluation reflects the model’s true generalization capability, free from biases introduced by data leakage, imbalanced sampling, or unintended shortcut learning.

\section{Full Experiment Results}

This appendix presents the complete experiment results for NSP and NAE, complementing the analyses in Section~\ref{sec:ablation} of the main text. The additional experiments here confirm the robustness of our hyperparameter choices and further validate the core insight that method‑specific features act as non‑transferable shortcuts.

\subsection{Detailed Efficiency Analysis}
\label{app:efficiency}

\begin{table*}[h]
\centering
\caption{Complete efficiency comparison on Xception backbone.}
\label{tab:efficiency_full}
\small
\renewcommand{\arraystretch}{0.5}
\resizebox{0.9\textwidth}{!}{\begin{tabular}{lcccc}
\toprule
\textbf{Method} & \textbf{Train time (s/epoch)} & \textbf{GPU mem (GB)} & \textbf{Infer time (ms/img)} & \textbf{Params} \\
\midrule
Xception (baseline) & 72 & 12.27 & 5.57 & 20.8M \\
Xception + NSP (Ours) & 72+30 & 12.27 & 5.64 & 20.8M + 18.5K \\
Xception + NAE (Ours) & 0+16 & 1.74 & 5.69 & 20.8M + 18.5K \\
\midrule
F$^3$Net & 76 & 12.99 & 5.99 & 21.4M \\
SPSL & 73 & 12.32 & 5.76 & 20.8M \\
SRM & — & >24 & 10.22 & 53.2M \\
RECCE & 410 & 19.20 & 50.43 & 23.8M \\
UCF & — & >24 & 9.12 & 46.8M \\
Deepspace & 73 & 12.29 & 5.86 & 22.0M \\
SpecXNet & 330 & 16.57 & 51.70 & 53.8M \\
\bottomrule
\end{tabular}}
\end{table*}

This section provides complete efficiency results omitted from the main text due to space constraints. All experiments are conducted on a single NVIDIA RTX 4090 GPU (24 GB memory) with batch size 64 for training and batch size 1 for inference. Training time is reported as seconds per epoch; inference time is milliseconds per image; peak GPU memory usage is in GB.

Table~\ref{tab:efficiency_full} lists all methods evaluated on the Xception backbone. Our methods (NSP and NAE) introduce negligible overhead while maintaining strong generalization. In contrast, several competing detectors either exceed memory capacity or incur significant latency.

\subsection{Complete NSP Ablation}

\begin{figure*}[ht]
  \centering 
  \includegraphics[width=0.92\textwidth]{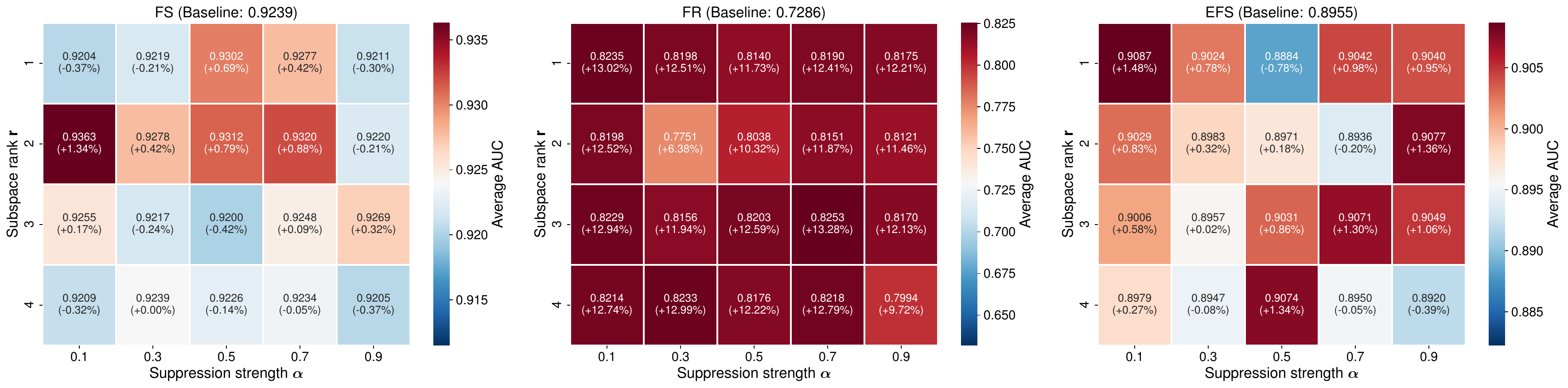}
  \caption{Complete NSP ablation. Each cell shows the average cross‑domain AUC across all test sets. The model is trained on the indicated domain (FS, FR, or EFS) and evaluated with varying subspace rank \(r\) and suppression strength \(\alpha\). The baseline performance (no NSP) is marked for reference.}
  \label{fig:nsp_heatmap} 
\end{figure*}

Figure~\ref{fig:nsp_heatmap} shows the full NSP ablation results across three training configurations (FS, FR, EFS) and all test sets. Each cell reports the average AUC across the four test categories.

Consistent with the main text, NSP mostly outperforms the baseline across all three training domains, though the gains depend on the choice of \(r\) and \(\alpha\). The optimal configuration varies slightly across training domains, but the region around \(r=3\) and \(\alpha=0.5\) generally yields the strongest cross‑domain generalization.

Notably, when training on the FR domain, the improvement is particularly pronounced: across all parameter combinations, NSP achieves at least a \(6.38\%\) absolute AUC gain over the baseline. This reflects the fact that the FR domain contains especially strong method‑specific shortcuts—models trained on FR heavily overfit to these patterns, leaving substantial room for improvement when we suppress them. The consistent gains across all \(r\) and \(\alpha\) choices further demonstrate that even suboptimal NSP configurations still effectively mitigate shortcut reliance in this challenging setting.

\subsection{Complete NAE Ablation}

\begin{figure*}[ht]
  \centering 
  \includegraphics[width=0.92\textwidth]{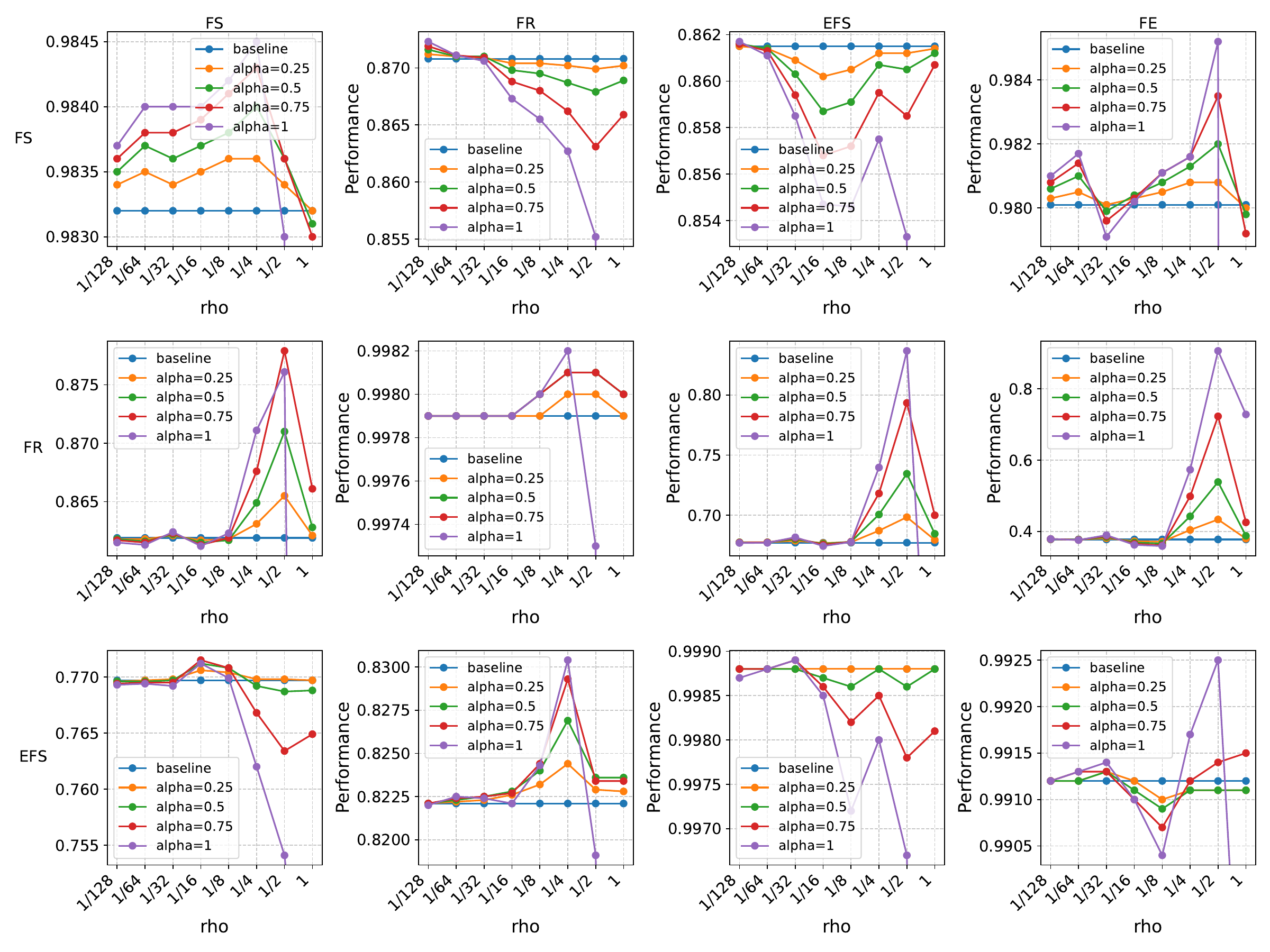}
  \caption{Complete NAE ablation. Each subplot shows cross‑domain AUC as a function of suppression ratio \(\rho\) (doubling from \(1/128\) to \(1\)) and strength \(\alpha\), averaged over test sets. The model is trained on the indicated domain (FS, FR, or EFS). Baseline performance (no suppression) is marked by the dashed line.}
  \label{fig:NAE_curve} 
\end{figure*}

Figure~\ref{fig:NAE_curve} presents the full NAE ablation results. We vary the suppression ratio \(\rho\) across eight values (from \(1/128\) to \(1\), increasing by a factor of 2 each step) and the suppression strength \(\alpha\) from \(0\) to \(1\) in increments of \(0.25\), evaluating on three training domains and four test sets. Each curve represents the average cross‑domain AUC for a given training domain.

Several observations emerge from this full sweep. First, the optimal suppression ratio \(\rho\) consistently falls within \([1/4, 1/2]\) across all training domains, aligning with the analysis in the main text. Second, performance improves monotonically with \(\alpha\): stronger suppression yields higher cross‑domain accuracy. This trend directly supports our core argument—method‑specific features act as non‑transferable shortcuts, and suppressing them more aggressively pushes the model toward more robust, generalizable cues.

The FR training domain again exhibits a distinctive pattern. For models trained on FR, even a small amount of suppression yields a sharp performance increase, far exceeding the gains observed on FS or EFS. This sharp rise reflects the severity of shortcut reliance in FR‑trained models: their representations are so dominated by method‑specific patterns that removing even a fraction of the associated neurons unlocks substantial generalization capacity. As \(\rho\) increases beyond the optimal range, performance eventually declines—a pattern consistent across all domains, indicating that overly aggressive suppression risks removing generalizable cues.

When \(\rho\) is too small (e.g., \(1/128\) or \(1/64\)), the model suppresses too few neurons, leaving method‑specific shortcuts largely intact; performance therefore remains close to the baseline. When \(\rho\) is too large (e.g., \(1\)), suppression extends to neurons that may carry generalizable forensic information, causing accuracy to drop. These patterns hold consistently across all three training domains, demonstrating that NAE behaves predictably and that the identified hyperparameter region generalizes across settings.

Together, the full ablation results confirm that both NSP and NAE improve generalization by suppressing method‑specific features, and that the hyperparameter choices reported in the main text are well‑justified across diverse training configurations.